\documentclass[10pt,journal,compsoc]{IEEEtran}
\ifCLASSOPTIONcompsoc
  \usepackage[nocompress]{cite}
\else
  \usepackage{cite}
\fi
\ifCLASSINFOpdf
  \usepackage[pdftex]{graphicx}
\else
  \usepackage[dvips]{graphicx}
\fi

\usepackage{amsmath}
\usepackage[linesnumbered,ruled,vlined]{algorithm2e}

\usepackage{algorithmic}

\usepackage{array}
\usepackage{multirow}

\usepackage{subcaption}

\usepackage{fixltx2e}

\usepackage{stfloats}
\ifCLASSOPTIONcaptionsoff
 \usepackage[nomarkers]{endfloat}
\let\MYoriglatexcaption\caption
\renewcommand{\caption}[2][\relax]{\MYoriglatexcaption[#2]{#2}}
\fi

\usepackage{url}

\usepackage[utf8]{inputenc} 

\usepackage{amssymb,amsthm}
\usepackage[dvipsnames]{xcolor}
\usepackage{soul}
\usepackage{caption}
\usepackage[]{changes}

\definechangesauthor[name={Michael}, color=blue]{MB}
\definechangesauthor[name={Luca},color=red]{LC}
\definechangesauthor[name={Anees Kazi},color=green]{AK}

\newcommand{\new}[1]{\color{black} #1 \color{black}}
\newcommand{\newtab}{\color{black}}

\usepackage[utf8]{inputenc}
\usepackage{pgfplots}
\usepgfplotslibrary{groupplots,dateplot}
\usetikzlibrary{patterns,shapes.arrows}
\pgfplotsset{compat=newest}

\begin{document}
%
\title{Differentiable Graph Module (DGM) for Graph Convolutional Networks}
%
%
%
%

\author{Anees~Kazi*,
        Luca~Cosmo*,
        Seyed-Ahmad Ahmadi,
        Nassir Navab,
        and~Michael M. Bronstein
\IEEEcompsocitemizethanks{
\IEEEcompsocthanksitem A. Kazi and N. Navab are with the department of Computer Aided Medical Procedures and Augmented Reality at Technical University of Munich. N. Navab is also with the Whiting School of Engineering, Johns Hopkins University, Baltimore, USA;
\IEEEcompsocthanksitem L. Cosmo is with the Ca' Foscari University of Venice, Italy, and with the Faculty of Informatics, University of Lugano, Switzerland;
\IEEEcompsocthanksitem S.-A. Ahmadi is with NVIDIA, Munich, Germany; 
\IEEEcompsocthanksitem M. M. Bronstein is with the department of Computer Science at the University of Oxford, UK and with Twitter, UK
}
\thanks{* A. Kazi and L. Cosmo  contributed equally to this work. Part of this work was done during A. Kazi's research visit at Imperial College London.\\ Corresponding author: A. Kazi }
} 

%
%

\markboth{Journal of \LaTeX\ Class Files,~Vol.~14, No.~8, August~2015}%
{Kazi \MakeLowercase{\textit{et al.}}:Differentiable Graph Module (DGM) for Graph Convolutional Networks}
%



\IEEEtitleabstractindextext{%

\begin{abstract}
Graph deep learning has recently emerged as a powerful ML concept allowing to generalize successful deep neural architectures to non-Euclidean structured data. Such methods have shown promising results on a broad spectrum of applications ranging from social science, biomedicine, and particle physics to computer vision, graphics, and chemistry. One of the limitations of the majority of current graph neural network architectures is that they are often restricted to the transductive setting and rely on the assumption that the underlying graph is {\em known} and {\em fixed}. Often, this assumption is not true since the graph may be noisy, or partially and even completely unknown. In such cases, it would be helpful to infer the graph directly from the data, especially in inductive settings where some nodes were not present in the graph at training time. Furthermore, learning a graph may become an end in itself, as the inferred structure may provide complementary insights next to the downstream task. In this paper, we introduce Differentiable Graph Module (DGM), a learnable function that predicts edge probabilities in the graph which are optimal for the downstream task. DGM can be combined with convolutional graph neural network layers and trained in an end-to-end fashion. We provide an extensive evaluation of applications from the domains of healthcare (disease prediction), brain imaging (age prediction), computer graphics (3D point cloud segmentation), and computer vision (zero-shot learning). We show that our model provides a significant improvement over baselines both in transductive and inductive settings and achieves state-of-the-art results.
\end{abstract}

\begin{IEEEkeywords}
Graph convolution, Graph learning, Disease prediction
\end{IEEEkeywords}}

\maketitle

\IEEEdisplaynontitleabstractindextext

%
\IEEEpeerreviewmaketitle

\IEEEraisesectionheading{\section{Introduction}\label{sec:introduction}}

\IEEEPARstart{G}{eometric} deep learning (GDL) is a novel emerging branch of deep learning attempting to generalize deep neural networks to non-Euclidean structured data such as graphs and manifolds \cite{bronstein2017geometric,hamilton2017representation,battaglia2018relational}. Graphs, being general abstract descriptions of relation and interaction systems, are ubiquitous in different branches of science. 
Graph-based learning models have been successfully applied in social sciences \cite{zhang2018link,qi2018learning}, computer vision and graphics \cite{qi20173d,monti2017geometric,wang2019dynamic}, physical  \cite{choma2018graph,duvenaud2015convolutional,gilmer2017neural,li2018learning}, as well as medical and biological sciences \cite{parisot2018disease,parisot2017spectral,mellema2019multiple,kazi2019inceptiongcn,zitnik2018modeling,Zitnik19,gainza2019deciphering}. 
%
%

Graph Neural Networks (GNNs) are a popular approach for learning on graphs. 
While dating back to at least \cite{scarselli2008graph}, it is mainly the recent progress that has made GNNs a useful and popular tool. 
Today's wide variety of GNN architectures includes spectral \cite{bruna2013spectral} and spectral-like \cite{defferrard2016convolutional,kipf2016semi,levie2018cayleynets,bianchi2019graph} methods, local charting \cite{monti2017geometric}, and attention \cite{velivckovic2017graph,kondor2018n,bruna2017community,monti2018dual}. 
Battaglia et al. \cite{battaglia2018relational} showed that most GNNs can be formulated in terms of message passing \cite{gilmer2017neural}. 
Recent efforts have been devoted to scalable 
\cite{hamilton2017inductive,pinsage,Chiang:2019:CEA:3292500.3330925,DBLP:journals/corr/abs-1907-04931} and deep \cite{rong2019dropedge,zhao2019pairnorm,li2019deepgcns,gong2020geometrically} GNN architectures, and theoretical analyses of their expressive power \cite{xu2018how,DBLP:conf/iclr/MaronBSL19,keriven2019universal,Loukas2020What}.  
%

%
%
%
%
%

A notable drawback of most GNN architectures is the assumption that the underlying graph is  {\em given} and {\em fixed}. Given this graph, convolution-like operations typically amount to modifying the node-wise features. 
Architectures like message passing neural networks \cite{gilmer2017neural} or primal-dual convolutions \cite{monti2018dual} also allow to update the edge features, but the graph {\em topology} is always kept the same. 
This often happens to be a limiting assumption. In many problems, the data can be assumed to have some underlying graph structure, however, the graph itself might not be explicitly given \cite{liu2012robust}, a setting we refer to as {\em latent graph}.  
In medical and healthcare applications, for example in patient population models, the graph between patients may be noisy or partially and even completely unknown, and one is thus interested in inferring it from the data. Such latent graphs can capture the actual topology of structured data, aiding towards efficient processing and analysis while simultaneously learning the relationship between the behavioral influence among the nodes. In fact, sometimes the graph may be even more important than the downstream task, as it conveys some interpretability of the model and may provide a means for knowledge discovery, e.g. in form of patient relations in a population model. 
Being able to learn a latent graph is especially important in case of inductive settings, 
where some nodes might be present in the graph at testing but not training time.

Graph topology inference is a long-standing problem and has recently been addressed using signal processing techniques \cite{Dong19,Mateos19}. We provide a detailed literature survey for different types of graph learning techniques in the next sections. 
In our recent work \cite{cosmo2020latent}, we proposed a graph learning model targeted towards node classification as the primary task. The main methodological contribution is the latent-graph learning block, which is capable of learning the probabilistic graph in form of a weighted adjacency matrix for the optimal classification outcome. 
The main limitation of this approach \cite{cosmo2020latent} was to model the graph in quasi-probabilistic and thus fully connected manner, since the model lacked the capability to exploit a possible sparsity of the graph. 
In this work, we propose a more general Differentiable Graph Module (DGM) for latent graph prediction. We propose both a continuous, generalizing \cite{cosmo2020latent}, and a discrete sampling strategy that overcomes the dense graph limitations and allows addressing bigger graphs. Further, our model explicitly models graph sparsity, resulting in a faster computation and lower memory consumption. 
%
\subsubsection*{Main contributions}
\begin{itemize}
    \item We propose an end-to-end pipeline for simultaneous learning of the downstream task, e.g node classification, along with an optimal underlying latent graph.
    \item We propose DGM with two sampling strategies, a continuous version that improves and generalizes our previously proposed method \cite{cosmo2020latent}, and a novel discrete version that can efficiently learn larger graphs.
    \item We show proof of concept experiments, on three benchmark graph datasets for node and graph classification. 
    \item We provide extensive ablation studies for both our model and discrete sampling strategy. Our evaluation demonstrates applications from several domains, i.e. healthcare and brain imaging (disease and age prediction), computer graphics (3D point cloud segmentation), and computer vision (zero-shot learning). Our model shows significant improvement over baselines and achieves state-of-the-art results. 
    
\end{itemize}

\new{
The Pytorch implementation of DGM can be found at the following Github repository: \url{https://github.com/lcosmo/DGM_pytorch}.}
\begin{figure*}[t]
\centering
  \includegraphics[width=1.05\textwidth]{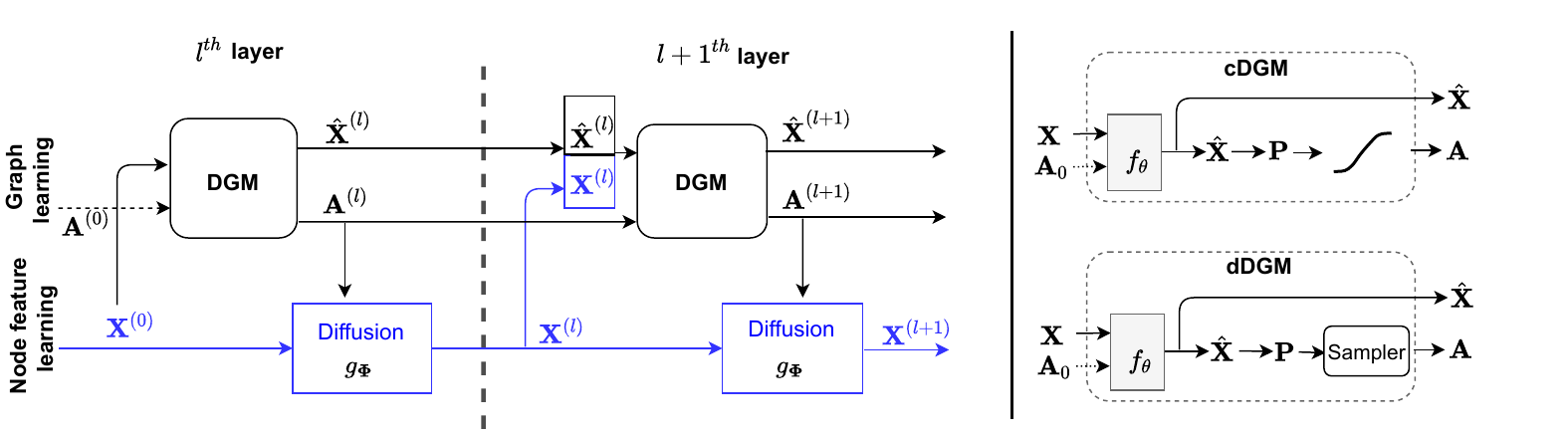}
  \caption{{\em Left:} Two-layered architecture including Differentiable Graph Module (DGM) that learns the graph, and Diffusion Module that uses the graph convolutional filters. {\em Right:} Details of DGM in its two variants, cDGM and dDGM.} \vspace{-3mm}
  \label{DGM}
\end{figure*}

\section{State of the art}
Recently, graph learning has come to focus due to its applicability in various domains \cite{dong2019learning}. In this section, we provide a detailed state of the art on graph learning, subdividing it into three main categories.

\paragraph*{\textbf{Attention-based graph learning}}
In the machine learning literature, 
several models dealing with latent graphs have recently been proposed. In this category, we include works that deal with attention to the nodes and edges, since 
edge-based attention can be seen as a latent structure learning process.
%
Kipf et al. \cite{kipf2018neural} used a variational autoencoder, in which the latent code represents the interaction graph underlying a physical system, and the reconstruction is based on graph neural networks. Further, graph attention networks (GAT) \cite{velivckovic2017graph} proposed an additional attention mechanism that operates on the edges of a fixed graph to learn the importance of each neighbor. 
Gated Attention Network (GAAN)\cite{zhang2018gaan} 
proposes a self-attention mechanism that computes an additional attention score for each attention head. RNN-based attention models have been proposed in  \cite{lee2018graph,abu2017watch,liu2019geniepath}.
The common drawback of these methods is that they need an input graph. Further, attention is computed directly on edge features, requiring the graph to be sparse in order to make the approach computationally feasible, especially for bigger graphs. 

\paragraph*{\textbf{Models dealing with dynamic graph}}
These methods do not assume a fixed graph a-priori, but build them dynamically during training. In their seminal work, Wang et al. \cite{wang2019dynamic} proposed Dynamic Graph CNNs (DGCNN) for the analysis of point clouds, where a k-nearest neighbor (kNN) graph is constructed on the fly in the feature space of the neural network. This idea is further extended in PGC-DGCNN \cite{tran2018filter}, which aims at increasing the contributions of distant neighbors using shortest paths connections.
In both of these methods \cite{wang2019dynamic, tran2018filter}, the latent space in which the graph is constructed is not optimized for the sought latent structure. Instead, they sample local neighborhoods of feature representations that are used for the downstream task. Importantly, the kNN operation used to build the graph is not directly differentiable.

%

\paragraph*{\textbf{Models dealing with graph learning}}
These methods focus on learning the graph directly. We subdivide this category into two groups. 

The first group operates in the spectral domain. Zhan et al. \cite{zhan2018adaptive} proposed to learn the latent structure by constructing multiple Laplacians and combining them with learnable weights. Similarly, Li et al. \cite{li2018adaptive} proposed a spectral graph convolutional method, in which a residual Laplacian computed on the feature output from each layer and the input Laplacian are updated after each layer. Huang et al. \cite{huang2018adaptive} proposed another version of spectral filters that parametrize the Laplacian instead of the coefficient of the filter. 
All the spectral approach based methods need an initial graph and suffer from the limitation of transductive models, i.e. that the validation and/or test set need to be embedded into the training graph.

The second group operates in the spatial domain. Jiang et al. \cite{jiang2019semi} proposed a model where graph learning and graph convolution are integrated in a unified network architecture.
Franceschi et al. \cite{franceschi2019learning} formulated graph learning as a bi-level optimization problem, by modeling the graph as a hyper-parameter and optimizing it with a separate loss. 
%
%
A different strategy was adopted by Yang et al. \cite{yang2019modeling}, who proposed a method for giving scores to the nodes during sub sampling, but without learning the graph. \new {Similarly, \cite{ying2018hierarchical} proposed DiffPool, a differentiable graph pooling module to generate hierarchical representations of graphs. The method learns a differentiable soft cluster assignment for nodes at each layer mapping nodes to a set of clusters. Each cluster then becomes a node for the next GNN layer.}
%
Norcliffe et al. \cite{norcliffe2018learning} specifically tailored their method for visual question answering. They construct a joint embedding of word and image features and apply top-k sampling on the embedding distances. Their method is a two-step graph learning pipeline, and requires a heavily customized optimization techniques.
Yu et al. \cite{chen2020iterative} proposed to learn a fully connected graph in an end-to-end pipeline, sparsifying the graph according to a threshold in the graph embedding space. While effective, this strategy does not provide any means to control the graph topology, and importantly, it does not prevent the graph structure from degenerating e.g. into many disconnected sub-graphs.
\new{
More recently, Cosmo et al. \cite{cosmo2021graph} proposed a graph convolution operator that directly learns both the connectivity and features of a set of graphs (i.e. structural masks) to be used as local filters on the input graph.
}

Compared to the methods above, we propose a technique for latent graph discovery which learns the graph end-to-end, which is inductive, scalable to large numbers of nodes, and which gives the network designer a means to control the graph topology. 


\section{Background}
Given a set of $N$ data points of dimension $d$, denoted $\textbf{X}\in\mathbb{R}^{N\times d}$, a common problem in machine learning is to produce a representation that is aware of the underlying structure of the data. 
Such structure can be represented as a (weighted) graph $\mathcal{G} = (\mathcal{V}, \mathbf{A})$ where $\mathcal{V} = \{1,\hdots, n\}$ is the vertex set and $\mathbf{A} = (a_{ij})$ is a (weighted) adjacency matrix. We use $\mathbf{A}$ to define the edges of the graph $\mathcal{E} = \{ (i,j) : a_{ij} > 0, \, i,j \in \mathcal{V} \}$;  the edge weight $a_{ij} \geq 0$ represents the affinity between points $\mathbf{x}_i$ and $\mathbf{x}_j$. 

\subsubsection{Graph neural networks}
Assuming this structure is provided together with the data, we have a node-attributed graph $\mathcal{G} = (\mathcal{V}, \mathbf{A}, \mathbf{X})$, on which a graph neural network (GNN) can be applied. 
GNN attempts to find an {\em embedding} $\mathbf{Z} = g_{\boldsymbol{\Theta}}(\mathbf{X}, \mathbf{A})$ by doing message passing \cite{gilmer2017neural,battaglia2018relational} 
\vspace{-2mm}
\begin{equation}
\mathbf{z}_i = \sum_{j \in \mathcal{N}_i} h_{\boldsymbol{\Theta}}(\mathbf{x}_i,\mathbf{x}_j, a_{ij}) 
\label{eq:ec}
\end{equation}
in a local neighborhood $\mathcal{N}_i = \{ j : (i,j) \in \mathcal{E} \}$ of the node. Here $h_{\boldsymbol{\Theta}}$ denotes a learnable function shared across nodes, whose parameters $\boldsymbol{\Theta}$ are chosen to minimize a downstream loss. 
Equation~(\ref{eq:ec}) is also referred to as {\em edge convolution} (EC) \cite{wang2019dynamic} due to its generalization of the classical convolution operation on grids. 
A particular case of~(\ref{eq:ec}) with node-wise linear transformation $h_{\boldsymbol{\Theta}} = a_{ij} \boldsymbol{\Theta}\mathbf{x}_j$ by matrix $\boldsymbol{\Theta}$, is called {\em graph convolution} (GC) \cite{defferrard2016convolutional,kipf2016semi}. 
The node embeddings can be used for node-wise classification, or pooled for graph-wise classification tasks. 
\subsubsection{Latent graphs}
We are interested in the setting when the underlying graph is {\em unknown}, and thus needs to be learned. 
Learning the graph serves two purposes: First, it is used to represent the structure of the data. Second, it is used as the support for graph-based convolutions to obtain the embeddings of the data points. 
%
%

The main obstacle for including the graph construction as a part of the deep learning pipeline is that it is a discrete structure and as such non-differentiable.     
%
%
To avoid this problem, in DGCNN \cite{wang2019dynamic}, the graph convolutional filters and the layer activations are optimised towards the downstream classification and segmentation tasks. The graph, however, is constructed ad-hoc as a kNN graph on the activations after each layer, without a dedicated loss. As such, the graph is built dynamically but not learned, and the underlying latent graph of the domain is not recovered.
 Our method, described in the next section, aims at addressing this issue. 

\section{Method}

%

%

\subsection{Architecture}

We propose a general technique for learning the graph based on the output features of each layer, along with the optimization of the network parameters during the training. 
Our architecture comprises two main blocks, the {\bf Differentiable Graph Module (DGM)} and the {\bf Diffusion Module}. \new{The interaction between these two modules is shown in Figure~\ref{DGM} and further detailed in Algorithm \ref{alg:algo}.} 

\subsubsection{Differentiable Graph Module:} The DGM is tasked with building the (weighted) graph representing the input space. 
It takes the feature matrix $\mathbf{X} \in \mathbb{R}^{N\times d}$ as input and yields a graph $\mathcal{G}$ as output. DGM can also take an initial graph $\mathcal{G}_0$ as input, in case an expert or domain knowledge exists allowing to define the initial edge weights between the nodes. Importantly, such an initial graph is optional, making DGM usable also in cases where domain knowledge is absent or where it is not desirable to impose a graph prior.
Since the node set $\mathcal{V}$ is fixed, the two graphs can be represented by their adjacency matrices, ${\bf A}_0$ and ${\bf A}$. 
\begin{figure*}[t]
    \centering
    \includegraphics[width=0.8\linewidth]{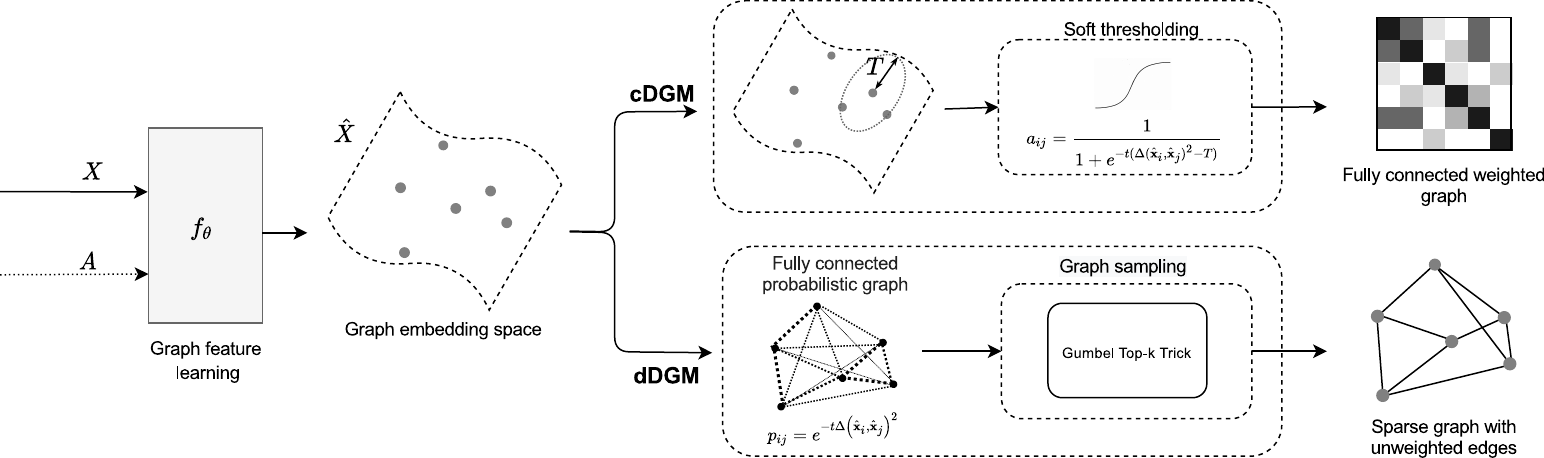}
    \caption{\newtab  Differentiable Graph Module (DGM) described in detail. After learning the lower dimensional embedding space the diagram shows proposed sampling techniques. Upper and lower part shows the continuous (cDGM) and discrete (dDGM) sampling variants.}
    \label{fig:DGM_module} 
\end{figure*}

\new{Figure \ref{fig:DGM_module} depicts the inner functioning of the Differentiable Graph Module.}
Input features $\mathbf{X} \in \mathbb{R}^{N\times d}$ are first transformed into {\em auxiliary features} $\hat{\textbf{X}}=f_{\boldsymbol{\Theta}} (\textbf{X}) \in \mathbb{R}^{N\times \hat{d}}$ by means of a parametric function $f_{\boldsymbol{\Theta}}$, typically reducing the input dimension ($\hat{d} \ll d$). If the initial graph $\mathcal{G}_0$ is provided, we can use $f_{\boldsymbol{\Theta}}$ of the general form~(\ref{eq:ec}), where new features $\hat{\textbf{X}}$ are computed by edge- or graph-convolutions on $\mathcal{G}_0$. 
Otherwise, $f_{\boldsymbol{\Theta}}$ is applied to each node feature independently, acting row-wise on the matrix $\mathbf{X}$. %

Second, the auxiliary features $\hat{\textbf{X}}$ are used for graph construction. We define the edge probabilities as:
\begin{equation}
p_{ij}(\mathbf{X};  \boldsymbol{\Theta},t) =  e^{-t \Delta\left ( \hat{\textbf{x}}_{i},\hat{\textbf{x}}_{j} \right )^2} = e^{-t \Delta \left ( f_{\boldsymbol{\Theta}} (\textbf{x}_i), f_{\boldsymbol{\Theta}} (\textbf{x}_j) \right )^2},
\label{eq:prob}
\end{equation}
where $t$ is a learnable parameter and $\Delta(\cdot,\cdot)$ is the distance between two nodes in the graph embedding space.
In the experimental session, we investigate the use of two different metric definitions, namely the Euclidean and the Hyperbolic metrics \cite{krioukov2010hyperbolic,poincare_embeddings}.
%

\vspace{0.5em}
\noindent\textbf{Continuous sampling:\hspace{0.5em}}
A straightforward way to derive a graph $\mathcal{G}$ is to transform the probability matrix $\mathbf{P}(\mathbf{X};\boldsymbol{\Theta},t)$ into a weighted adjacency matrix using the sigmoid function 
%
 $a_{ij}=1/(1+p_{ij} e^{tT})$.
%
\new{Here, as depicted in figure \ref{fig:DGM_module}, $T$ can be interpreted as a threshold on the squared distances $\Delta( \hat{\textbf{x}}_{i}-\hat{\textbf{x}}_{j} )^2$ defined on the embedding space. Indeed, elements $a_{ij}$ can be rewritten as $a_{ij} = \frac{1}{1+ e^{-t(\Delta (\hat{\mathbf{x}}_{i}, \hat{\mathbf{x}}_{j})^{2} - T)}}$.}
In this way, the graph is represented by the adjacency matrix $\mathbf{A}(\mathbf{X};\boldsymbol{\Theta},t,T)$ parametrized through $\boldsymbol{\Theta}, t$ and the additional parameter $T$, and is differentiable w.r.t. these parameters.  

This sampling strategy generalizes the method proposed in \cite{cosmo2020latent}. We refer on this paper to this variant as {\em continuous DGM (cDGM)}.   

\vspace{0.5em}
\noindent\textbf{Discrete sampling:\hspace{0.5em}}
One of the shortcomings of cDGM is that it can produce a dense adjacency matrix, i.e. a fully connected graph in which many edges have near-zero weight. 
As an efficient alternative, we propose to construct a sparse $k$-degree graph by using the \textit{Gumbel-Top-$k$ trick} \cite{wouter2019} to sample edges from the probability $\mathbf{P}(\mathbf{X};\boldsymbol{\Theta},t)$. 
Such sampling can be regarded as a stochastic relaxation of the kNN rule. 
%
%
%
%
For each node $i$, we extract $k$ edges $(i,j_{i,1}), \hdots, (i,j_{i,k})$ as the first $k$ elements of 
\begin{equation}
\mathrm{argsort} (\log(\textbf{p}_i) -\log(-\log(\textbf{q})),
\end{equation}
where $\textbf{q}\in\mathbb{R}^N$ is uniform i.i.d. in the interval $[0,1]$.
Samples extracted this way follow the categorical distribution $p_{ij}/ \sum_r p_{ir}$  \cite{wouter2019}. 
We define the edge set of the sparse graph $\mathcal{G}$ constructed this way as
\begin{equation}
\mathcal{E}(\mathbf{X};\boldsymbol{\Theta},t) = \{ (i,j_{i,1}), \hdots, (i,j_{i,k}) : i = 1, \hdots, N \}    
\end{equation}
and represent it by the unweighted adjacency matrix $\mathbf{A}(\mathbf{X};\boldsymbol{\Theta},t)$. The key advantage is that this matrix is sparse, which results in a lower computational and memory complexity of the diffusion operation.
%
%
%
We refer to this variant of our architecture as \textbf{{\em discrete DGM (dDGM)}}.   

Note that, since the dDGM graph sampling scheme is stochastic, the prediction of the network at inference time is not deterministic. We can actually take advantage of this, and implement a consensus scheme. In our experiments, we run the classification for 8 times and select the maximum of the cumulative soft predictions as the predicted class. 

\subsubsection{Diffusion Module:} This module takes the graph affinity $\mathbf{A}$ produced by the DGM and the features $\mathbf{X}^{(l)}$ as inputs, and yields a new set of features $\mathbf{X}^{(l+1)} = g_{\boldsymbol{\Phi}}(\mathbf{X}^{(l)})$ as output as shown in Fig \ref{DGM}. Here, $g_{\boldsymbol{\Phi}}$ represents a general function of the form~(\ref{eq:ec}); in our experiments, it is either edge- or graph-convolution on $\mathbf{A}$. 
\subsection{Classification model (Combined model):} 
We use a multi-layer network, with layers numbered as $l=1,\hdots, L$.
Each layer comprises a {\bf DGM} and {\bf Diffusion Module}, as shown in Figure~\ref{DGM}.
%
The $l$th layer of the architecture 
%
produces the output as: 
\begin{align}
     \hat{\textbf{X}}^{(l+1)} &= f^{(l+1)}_{\boldsymbol{\Theta}} ([\textbf{X}^{(l)} \, | \, \hat{\mathbf{X}}^{(l)}], \mathbf{A}^{(l)}) \quad\quad\\
     {\textbf{A}}^{(l+1)} &\sim \textbf{P}^{(l)}(\hat{\textbf{X}}^{(l+1)}) \quad\quad\\ 
     \textbf{X}^{(l+1)} &= g_{\boldsymbol{\Phi}} (\mathbf{A}^{(l+1)},\textbf{X}^{(l)})
\end{align}
We assume $\mathbf{X}^{(0)} = \mathbf{X}$ and unless some initial knowledge of the structure of the data is available, $\mathbf{A}^{(0)}=\mathbf{I}$ (i.e., the initial graph is $\mathcal{G}^{(0)} = (\mathcal{V},\emptyset)$) and $f^{(0)}_{\boldsymbol{\Theta}}$ is a node-wise function (MLP).   
In case of classification task, the final node features $\textbf{X}^{(L)}$ of the last layer $L$ can then be given as input to a MLP to obtain the final node predictions. Such an end-to-end model can be defined as $\hat{Y}=\mathbb{M}\left ( X^{0}, A^{0} \right )$
where, $\hat{Y}$ is predicted label vector.  

Note that DGCNN can be obtained as a particular setting of our model with  $f_{\boldsymbol{\Theta}} = \mathrm{id}$ in the {\bf DGM} module and using edge convolution in the {\bf Diffusion} module. \new{Here $\mathrm{id}$ stands for the identity mapping.}



%
%
\subsubsection{Loss function and Training}
To train end-to-end our cDGM model we optimize directly over the cross-entropy loss of the final classification task. However, the sampling scheme we adopt in dDGM does not allow the gradient of the downstream classification loss function to flow through the graph prediction branch of our network, as it involves only graph features $\mathbf{\hat{X}}$.

To allow its optimization, we 
create a compound loss that rewards edges involved in a correct classification and penalizes edges that led to misclassification. 

Let $\mathbf{y}$ denote the vector of node-wise labels predicted by our model at step $(t)$ and $\tilde{\mathbf{y}}$ the groundtruth labels.  

We define the reward function $\delta(y_i,\tilde{y}_i) = \mathbb{E}(\mathbf(a_i)) - a_i$ as the difference between the average accuracy of the $i$th sample and the current success value $a_i = 1$ if $y_i = \tilde{y}_i$ and $0$ otherwise.
We then derive the graph loss as:
%
\begin{equation}
    L_\mathrm{graph}(\boldsymbol{\Theta}^{(1)}, \hdots, \boldsymbol{\Theta}^{(L)}) = \sum_{\substack{i=1\dots N\\l=1\dots L\\j:(i,j)\in \mathcal{E}^{(l)}}} \delta(y_i,\tilde{y}_i) \, \log \, p_{ij}^{(l)}(\boldsymbol{\Theta}^{(l)})
    \label{eq:graph_loss}
\end{equation}
whose gradient approximates the gradient of the expectation 
$\mathbb{E}_{ (\mathcal{G}^{(1)}, \hdots, \mathcal{G}^{(L)})  \sim (\mathbf{P}(\boldsymbol{\Theta}^{(1)}),\hdots, \mathbf{P}(\boldsymbol{\Theta}^{(L)}) )} \sum_i \delta(y_i,\tilde{y}_i) $
with respect to the parameters of the graphs in all the layers.

We estimate $\mathbb{E}(\mathbf{a}_i)$ with two different strategies depending on the application. When the node set remains fixed for each sample and during training, we adopt an exponential moving average strategy on the accuracy values during the training process. In this case, $\mathbb{E}(a_i)^{(t+1)} = \alpha \mathbb{E}(a_i)^{(t)} + (1-\alpha) a_i$, with $\alpha=0.9$ in all our experiments. If the node set changes during training or between different samples, we estimate $\mathbb{E}(\mathbf{a}_i)$ individually for each training step $(t)$ evaluating the model multiple times on different sampled graphs.

\begin{algorithm}[t]
 \newtab
 \SetKwFunction{FDGM}{DGM}
 \KwData{$\mathbf{X}, \mathbf{A}$}
 \KwResult{$\hat{\mathbf{Y}}$}
 \Begin{
 $\mathbf{X}^{(0)} \gets \mathbf{X}$\\
 $\mathbf{A}^{(0)} \gets \mathbf{A}$\\
 $ \hat{\mathbf{X}}^{(0)} \gets \emptyset$\\
 \For{$x \in 0 \dots  L-1$}{
  $\hat{\mathbf{X}}^{(l+1)}, \textbf{A}^{(l+1)} \gets  \text{\FDGM}
  ([\mathbf{X}^{(l)} \, | \, \hat{\mathbf{X}}^{(l)}], \mathbf{A}^{(l)},f_{\boldsymbol{\Theta}}^{(l)},t^{(l)},T^{(l)})$\\
  \vspace{1.5mm}
  $\mathbf{X}^{(l+1)} \gets g_{\boldsymbol{\Phi}}^{(l)}(\mathbf{A}^{(l+1)},\mathbf{X}^{(l)})$\\
 }
 $\hat{\textbf{Y}} \gets MLP_{\Psi}(\textbf{X}^{(L)})$
 }
\hfill\\
\SetKwFunction{FDGMl}{DGM}
\SetKwProg{Fn}{Function}{:}{}
\Fn{\FDGMl{$\mathbf{X}$,$\mathbf{A},f_{\boldsymbol{\Theta}},t,T$}}{
    $\hat{\textbf{X}} \gets f_{\boldsymbol{\Theta}}(\textbf{X})$\\
    $p_{ij} \gets e^{-t \Delta\left ( \hat{\textbf{X}}_{i},\hat{\textbf{X}}_{j} \right )^2} $\\
    \SetKwSwitch{Switch}{case}{Other}{Switch}{:}{Case}{Other}{EndCase}{EndSwitch}
    \vspace{1mm}
    \Switch{sampling}{
    \Case{\textit{continuous}}{
        \For{$i,j \in 1 \dots N$}{
        $ a_{ij} \gets 1/(1+p_{ij} e^{tT})$
        }
    }
    \Case{\textit{discrete}}{
    \For{$i \in 1 \dots N$}{
        $\mathbf{q} \sim U(0,1)$\\
        $\mathbf{j_{\{k\}}}=\mathrm{argtopK} (\log(\mathbf{p_i})
        -\log(-\log(\mathbf{q}))$\\
        $
        a_{ij} = 
        \begin{cases}
    1,& j \in \mathbf{j_{\{k\}}}\\
    0,              & \text{otherwise}
    \end{cases}
        $
        }
        }
    }
    \KwRet $\hat{\textbf{X}}, (a_{ij})$\\
}
 \caption{Algorithm for the full model that produces $\hat{\textbf{Y}}$ i.e, the predicted labels which are then used by the loss function for training the model. With $(a_{ij})$ we refer to the matrix $\mathbf{A}$ composed by elements $a_{ij}$, while $\mathbf{p_i}$ refers to the $i$\textit{th} row of $P$. $ f_{\boldsymbol{\Theta}}$ and $g_{\boldsymbol{\Phi}}$  are generic function with learnable parameters $\Theta$ and $\Phi$ respectively, MLP is the Multi Layer Perceptron used for classification with parameters $\Psi$.}
 \label{alg:algo}
\end{algorithm}

\section{Experiments and Results}
We start the experimental session by showing ablation study results on popular benchmark datasets from \cite{planetoid} (CiteSeer, Cora, Pubmed). We then validate our approach against state-of-the-art  methods on the domains of healthcare applications (disease and age prediction) and show experiments on computer graphics (3D point cloud segmentation), and computer vision (zero-shot learning).


\subsection{Benchmark graph based datasets}
%
In this section we show the ablation study with respect to different hyper-parameters and architectural choices of our method to better analyze and evaluate the discrete sampling strategy proposed in this work (dDGM).
We make use of three popular citation graph datasets\cite{planetoid} for this task, namely Cora, PuMed and CiteSeer. Each of these datasets consists of a single graph in which nodes correspond to documents and edges to citation links. The goal is to predict the category of each document  transductively, i.e. the category of some documents is not known during training. Further details on these datasets are given in Table~\ref{tab:benchmark data_description}.

Since in these datasets the input graph is given, we take advantage of it by using a Graph Convolutional Layer (GCN) \cite{kipf2016semi} as the graph embedding function $f_{\Theta}$. Note that, even if the DGM module uses the input graph, the Diffusion module will have access only to the predicted graph (in Figure \ref{DGM} the input and the predicted graphs are represented by $\mathbf{A}^{(0)}$ and $\mathbf{A}^{(l)}$ respectively).

\textbf{Architecture details:} The base architecture for this ablation study is composed by one DGM layer where the graph is given as input to the associated diffusion layer. Its output is passed through a final MLP of size $(8,8,c)$, which classifies each node into one of $c$ classes. The DGM layer function $f$ is composed by two GCN layers with output feature dimensions of 16 and $d$ respectively. We set $d=4$ as the graph embedding space dimension. The diffusion function $g$ is composed of three GCN layers with output sizes 32, 16 and 8, respectively. We sample the kNN graph with a number of $k=5$ neighbors. \new{Before being processed by the DGM layer, input features are transformed by a perceptron (a linear transformation followed by a ReLU activation) of output size equal to 32.}
We train each model 10 times for 100 epochs each, using Adam optimizer with a learning-rate of 1e-2, and report the mean and standard deviation of accuracy (in percent).


\begin{table}[t]
    \caption{Dataset description of Cora, Citeceer and Pubmed }
    \label{tab:benchmark data_description}

    \centering
    \begin{tabular}{l| c| c| c| c}
\hline
 Dataset & nodes & edges & features & classes\\
\hline
Cora & 2,708 & 5,429 & 1,433 & 7\\
CiteSeer & 3,327 & 4,372 & 3,703 &6\\
Pubmed & 19,717 & 44,338 & 500 & 3\\
\hline
    \end{tabular}
\end{table}

\subsubsection{Latent graph sparsity}
In this experiment we consider the sparsity of the graph induced by the choice of the parameter $k$ while building the kNN graph.
The motivation behind this experiment is to evaluate the graph learning module at different sparsity levels and find the generic range of sparsity in terms of 'k'.
As shown in Table \ref{tab:k validation} the value of this parameter can have a significant impact on the performance of the network. In all the considered datasets a good range for $k$ is between 3 and 5. This means that the relations within the latent network can be captured by a highly sparse graph, which in turn leads to a high efficiency of our discrete sampling strategy.
\begin{table}[t!]
    \caption{Performance of our method with different numbers of $k$ nearest neighbors  when sampling the graph. We report the mean and standard deviation (in parenthesis) of the accuracy on 10 runs.}
    \label{tab:k validation}

    \centering
    \begin{tabular}{l| c| c| c}
\hline
 $k$ &Cora &PubMed&CiteSeer\\
\hline
1 & 72.80 (2.71e+0)&\textbf{88.60 (1.16e+0)}&67.00 (3.55e+0)\\
3 & 83.60 (9.91e-1)&\textbf{88.60 (8.85e-1)}&73.20 (1.49e+0)\\
5 & \textbf{84.60 (8.52e-1)}&87.60 (7.51e-1)&\textbf{74.80 (9.24e-1)}\\
10 & 81.80 (1.94e+0)&85.80 (1.71e+0)&72.00 (1.06e+0)\\
20 &70.60 (4.05e+0)&81.60 (1.38e+0)&64.80 (3.46e+0)\\
\hline

    \end{tabular}
\end{table}

\subsubsection{Graph embedding space geometry}
\label{sec:GES}
Recent literature in neural networks \cite{smith2019geometry} suggest that the geometry of the latent space in which data samples are embedded plays an important role in capturing the underlying relations. 
A $d$-dimensional hyperbolic space is a homogeneous space with constant negative curvature. 
For instance, hyperbolic spaces have been shown to better capture hierarchical structures compared to the standard Euclidean space \cite{chami2019hyperbolic}. With this experiment, we aim at investigating whether or not embedding the graph nodes in a hyperbolic space is beneficial for node classification. 

We model it using the Poincaré ball \cite{beltrami1868teoria}, where the distance between two points inside the unitary open ball, $x,y \in \mathcal{B}^d$, is computed as:
\begin{equation}
    d(\boldsymbol{x}, \boldsymbol{y})=\cosh ^{-1}\left(1+2 \frac{\|\boldsymbol{x}-\boldsymbol{y}\|^{2}_2}{\left(1-\|\boldsymbol{x}\|^{2}_2\right)\left(1-\|\boldsymbol{y}\|^{2}_2\right)}\right)\, ,
\end{equation}
where $\|.\|_2^2$ is the standard euclidean squared norm.  
To constrain the graph embedding function to predict values within the unit ball we apply a non linearity which clips the norm of the latent vectors to $1-\epsilon$.

In case of the Euclidean latent space geometry, we simply define the distance as $d(x,y)=\|x-y\|_2$.

Table \ref{tab:embb_space_m} shows the performance of models that learned a graph under the assumption of a hyperbolic vs Euclidean latent space.
For dDGM, as can be seen, the model with Euclidean embedding performs consistently better on the three investigated datasets. Further, the standard deviation shows that Euclidean space is more stable compared to the hyperbolic, indicating that the considered datasets do not draw any advantage from a hierarchical representation of node relations.

\begin{table}[t!]
    \caption{Performance of our method with euclidean and hyperbolic space geometries. We report the mean and standard deviation (within parenthesis) of the accuracy on 10 runs.
    }
    \label{tab:embb_space_m}
    \centering
    \begin{tabular}{l| c | c| c}
\hline
  &Cora &PubMed&CiteSeer\\
\hline
Euclidean& \textbf{84.60 (8.52e-01)} & \textbf{87.60 (7.51e-01)} & \textbf{74.80 (9.24e-01)}\\
Hyperbolic & 84.40 (1.70e+00) & 86.60 (9.52e-01) & 74.60 (7.63e-01)\\
\hline
    \end{tabular}
\end{table}


\subsubsection{Graph embedding space dimension}
The graph embedding dimension directly determines the subset of graphs that the DGM module is able to represent. A higher dimension allows to represent more complex relations between the nodes, while a smaller embedding space has a regularization effect that prevents overfitting.
The accuracy for different graph embedding space dimensions is shown in table \ref{tab:embb_space}. For the smaller datasets, Cora and CiteSeer, the best accuracy is obtained with a dimension of 4, while PubMed requires a higher embedding dimension of 8. The optimal value for the embedding space dimension thus depends on the complexity of the dataset. 

\begin{table}[t]
    \caption{Performance of our method with different graph embedding space dimensions $d$. We report the mean and standard deviation (within parenthesis) of the accuracy on 10 runs.}
    \label{tab:embb_space}
    \centering
    \begin{tabular}{l| c| c| c}
\hline
$d$&Cora&PubMed&CiteSeer\\
 \hline
2 & 80.20 (5.22e-01) & 87.80 (5.31e-01) & 71.40 (1.08e+00)\\
4 & \textbf{84.60 (8.52e-01)} & 87.60 (7.51e-01) & \textbf{74.80 (9.24e-01)}\\
6 & \textbf{84.60 (5.12e-01)} & 88.20 (6.01e-01) & 74.00 (6.76e-01)\\
8 & 84.40 (1.56e+00) & \textbf{88.40 (7.74e-01)} & 73.20 (1.20e+00)\\
\hline
    \end{tabular}
\end{table}

\subsubsection{Number of DGM Layers}
In this experiment, we investigate the impact of using the proposed DGM module rather than just performing graph convolutions on the input graph, and whether there is any benefit in stacking multiple DGM layers, i.e learning a dynamic graph.
In table \ref{tab:input_vs_no_input}, we report the accuracy with respect to the number of DGM layers. Here, a value of "0" denotes that we do not use DGM and perform the convolution directly on the input graph, corresponding to a conventional GCN based model. In practice, when using $L>0$ DGM modules, the diffusion module of the intermediate layers is composed by just one convolution layer, while the final diffusion layer will have $4-L$ convolution layers. This keeps the total number of convolution layers on the diffusion part equal to three.

Our results indicate that using a single graph, i.e a single DGM, is sufficient for increasing the classification accuracy w.r.t. the input graph in all datasets. Only PubMed benefits from a second DGM layer, again implying that it exhibits more complex relations between its nodes.

We remark that our method does not require an a-priori graph as input. If given, however, our results indicate that the graph embedding function $f$ can potentially exploit complementary information in its embedding, e.g. domain knowledge given by experts.


\begin{table}[t]
    \caption{Performance of our method with a different number of DGM layers. Zero layers means that DGM is not used. We report the mean and standard deviation (within parenthesis) of the accuracy on 10 runs.
    \label{tab:input_vs_no_input}}

    \centering
    \begin{tabular}{l| c| c| c}
\hline
&Cora&PubMed&CiteSeer\\
\hline
0 & 80.90 (1.19e+00)        & 86.40 (1.37e+00)  & 71.90 (1.50e+00)\\
1& \textbf{84.70 (1.40e+00)}& 87.60 (7.94e-01)  & \textbf{74.70 (1.18e+00)}\\
2& 83.90 (1.77e+00)         & \textbf{88.50 (5.59e-01)}& 74.50 (1.36e+00)\\
3& 82.40 (1.26e+00).        & 88.40 (5.37e-01)  & 72.90 (1.77e+00)\\
     \hline
    \end{tabular}
\end{table}

\new{
\subsubsection{Graph embedding function $f$}
We further show analysis of different graph embedding functions $f$ within DGM module. Table \ref{tab:arch_conv_f} shows the classification accuracy and its standard deviation using four different embedding functions $f$ on the 3 datasets. GCN \cite{kipf2016semi} and GAT\cite{velivckovic2017graph} functions make use of both node features and the input graph connectivity ($A_0$ in Figure \ref{DGM}). On the other hand, MLP applies nodewise a multilayer perceptron directly on the input features, discarding the input graph connectivity. To keep the number of parameters comparable, the MLP is composed by a linear layer of the same input and output dimensions of the convolution operations followed by a ReLU activation.
Finally, $\mathrm{id}$ represents the identity mapping, which builds a knn graph directly on the input feature space without any transformation.

Results show that using the input connectivity graph in the DGM module is essential to obtain reasonable performance, especially for Cora and CiteSeer datasets. Indeed, discarding the input graph connectivity by building \textit{de novo} the graph based on input features similarities (knn), or not using any connectivity information at all (MLP), leads to a huge drop ($10-20\%$) in accuracy. This means that, in these datasets, the graph structure itself conceive important information about the class each node belongs to.  
}
\begin{table}[t]
    \newtab
    \caption{\newtab Performance of our method with different graph embedding functions $f$. We report the mean and standard deviation (within parenthesis) of the accuracy on 10 runs.}
    \label{tab:arch_conv_f}

    \centering
    \begin{tabular}{l| c| c| c}
    \hline
 &Cora&PubMed&CiteSeer\\
  \hline
     gcn & 84.70 (1.40e-01) & \textbf{87.60 (7.94e-01)}& 74.70 (1.18e+00)\\
     gat & \textbf{84.90 (1.47e+00)}& 86.80 (9.14e-01) & \textbf{74.80 (1.17e+00)}\\
     mlp & 63.30 (2.86e+00)& 87.50 (8.00e-01)&64.90 (3.18e+00)\\
     id & 61.80 (4.11e+00)&84.20 (1.65e-01)& 59.90 (3.56e+00)\\
 \hline
    \end{tabular}
\end{table}

\subsubsection{Diffusion function $g$}
Here, we check the behaviour of the end-to-end pipeline with respect to different choices for the diffusion function $g$. We use three different convolutional layers with different characteristics: GCN \cite{kipf2016semi} performs a simple diffusion operation over the node features; GAT \cite{velivckovic2017graph} further learns a per-edge weight based on the attention mechanism; finally, EdgeConv operates over edge features to compute the output node feature.
As can be seen in table \ref{tab:arch_conv}, GCN is more stable and generally obtains the best performance, while using EdgeConv results in a comparably large drop of accuracy, especially on smaller datasets. A potential explanation is that the graph predicted by DGM is already optimal for the task at hand, and the edge weighting performed by GAT and intrinsically by EdgeConv does not give any contribution. Moreover, the graph sampled by DGM is different every time it gets sampled, and the resulting changes in the loss landscape may hinder convergence. In consequence, learning edge-related properties becomes a more challenging task, which can lead to a lower accuracy of GAT and EdgeConv.

\new{Further, we investigate the advantage given by using our DGM with respect to standard graph convolution neural networks. Table \ref{tab:arch_conv} shows the experiments with DGM i.e. performing the diffusion on the graph being learned, while table \ref{tab:arch_conv_b} without DGM, i.e. performing the diffusion directly on the graph given as input. As can be seen, irrespective of the specific diffusion function $g$, in most of the cases the model with DGM performs better than using just the input graph. Specifically, edgeconv shows worst performance as a diffusion model. The model with DGM and edgeconv performs up to $3.4\%$ better than without DGM. This experiment clearly shows the ability of our DGM to predict a graph more suitable for the task at hand, even when a predefined structure is already given as input to the classification model. Note that, both architectures with and without DGM have exactly the same number of parameters involved in the diffusion step and final classification MLP ($\approx 50K$ for the base architecture in Cora), while DGM needs to optimize some few extra parameters ($\le 150$ in the considered experiment) dedicated to the graph prediction. }


\begin{table}[t]
  \newtab
    \caption{ (a) Performance of our method with different diffusion functions $g$ on the graph predicted by DGM. (b) Same diffusion functions applied directly on the input graph (without DGM). We report the mean and standard deviation (within parenthesis) of the accuracy on 10 runs. }
    \centering
    \begin{subtable}[h]{1\linewidth}
    \caption{with DGM\vspace{-0.5em}}
    \label{tab:arch_conv}
    \begin{tabular}{l| c| c| c}
    \hline
 &Cora&PubMed&CiteSeer\\
  \hline
     gcn &\textbf{84.60 (8.52e-01)}&87.60 (7.51e-01)&\textbf{ 74.80 (9.24e-01)}\\
     gat & \newtab 81.30 (2.71e+00)& \newtab \textbf{87.80 (8.41e-01)}& \newtab74.00 (1.32e+00)\\
edgeconv &71.60 (4.20e+00)&87.40 (5.74e-01)& 52.20 (2.39e+00)\\
\hline
\multicolumn{4}{c}{}
\end{tabular}
\end{subtable}
\begin{subtable}[h]{1\linewidth}
\caption{without DGM\vspace{-0.5em}}
\label{tab:arch_conv_b}
\begin{tabular}{l| c| c| c}
 \hline
 &Cora&PubMed&CiteSeer\\
 \hline
     gcn & {80.90 (1.19e+00)} & \textbf{86.70 (1.37e+00)} & \textbf{ 71.90 (1.50e+00)}\\
     gat & \textbf{81.10 (1.79e+00)}& {84.60 (1.26e+00)}&71.20 (1.66e+00)\\
     edgeconv & 67.00 (3.94e+00) & 85.00 (1.54e+00)& 49.00 (4.52e+00)\\
 \hline
    \end{tabular}
    \end{subtable}
\end{table}

\subsubsection{Time and memory requirements.}
The key advantage of the proposed kNN discrete graph sampling strategy (dDGM) is the possibility to obtain a sparse graph, which makes graph convolutional operations more efficient in computational complexity and memory requirements.

In Figure \ref{fig:speedmemory}, we report both the time required for each training iteration and the allocated memory for three versions of our model depending on input number of nodes $n$. In this experiment, node features are randomly generated and the input graph is not considered, and the DGM embedding function $f$ is thus replaced with an MLP. cDGM refers to the continuous sampling strategy generalizing \cite{cosmo2020latent}. It is the most demanding implementation, since graph convolution is performed through dense matrix multiplication. With the kNN discrete graph sampling strategy (dDGM), graph convolutions operate on a sparse graph representation, resulting in lower memory consumption and faster computation, particularly on larger graphs with $n \gtrapprox 2000$. Still, the pairwise distance in the graph embedding space is performed through standard PyTorch operations, resulting in a quadratic time and memory complexity. In order to overcome this quadratic complexity as well, we provide a further, symbolic implementation using the KeOps library \cite{charlier2020kernel}. Using KeOps, the kNN sampling computes the pairwise distances on-the-fly instead of storing them in memory, resulting in a linear memory complexity and a much faster computation.


\begin{figure}[t]
    \centering
    \begin{minipage}{.49\linewidth}
\begin{tikzpicture}[scale=1]

\definecolor{color0}{rgb}{0.12156862745098,0.466666666666667,0.705882352941177}
\definecolor{color1}{rgb}{1,0.498039215686275,0.0549019607843137}
\definecolor{color2}{rgb}{0.172549019607843,0.627450980392157,0.172549019607843}
\scriptsize
\begin{axis}[
width=1.15\linewidth,
title={Training iteration time (s)},
legend cell align={left},
legend style={
  fill opacity=1,
  draw opacity=1,
  text opacity=1,
  at={(0.03,0.97)},
  anchor=north west,
  draw=white!80!black
},
tick align=outside,
tick pos=left,
xmin=500, xmax=9500,
xtick style={color=black},
y tick label style={
    /pgf/number format/.cd,
    fixed,
    fixed zerofill,
    precision=2
  },
ymin=-0.00111658334732056, ymax=0.255401346683502,
ytick style={color=black}
]
\addplot [line width=2pt, color0]
table {%
500 0.0139537811279297
1000 0.0105433225631714
1500 0.0116708755493164
2000 0.0151128053665161
2500 0.0209045886993408
3000 0.0273679971694946
3500 0.0353816032409668
4000 0.043990421295166
4500 0.0543893098831177
5000 0.0665093421936035
5500 0.0793586015701294
6000 0.0934281349182129
6500 0.107972979545593
7000 0.125397443771362
7500 0.14461510181427
8000 0.162740421295166
8500 0.184819340705872
9000 0.207339692115784
9500 0.24374144077301
};
\addlegendentry{cDGM}
\addplot [line width=2pt, color1]
table {%
500 0.0114662885665894
1000 0.0116960763931274
1500 0.0137805223464966
2000 0.0149441957473755
2500 0.0169662475585938
3000 0.020952296257019
3500 0.0242322444915771
4000 0.0273267269134521
4500 0.0342395782470703
5000 0.0398068428039551
5500 0.0461586236953735
6000 0.0533159255981445
6500 0.0602699995040894
7000 0.0687520980834961
7500 0.077764105796814
8000 0.0873364925384521
8500 0.0982192516326904
9000 0.109455323219299
9500 0.122003817558289
};
\addlegendentry{dDGM}
\addplot [line width=2pt, color2]
table {%
500 0.0177811145782471
1000 0.0171787023544312
1500 0.0170603275299072
2000 0.0174545764923096
2500 0.0163631916046143
3000 0.0178829669952393
3500 0.0161146879196167
4000 0.0161866903305054
4500 0.0180084705352783
5000 0.0186122417449951
5500 0.0173924446105957
6000 0.0185589790344238
6500 0.0176516771316528
7000 0.0184558868408203
7500 0.0180613994598389
8000 0.0184919357299805
8500 0.0185047388076782
9000 0.0179198026657105
9500 0.0190580129623413
};
\addlegendentry{dDGM Keops}
\end{axis}

\end{tikzpicture}
        \vspace{-1.5em}
    \end{minipage}
    \begin{minipage}{.49\linewidth}
\begin{tikzpicture}

\definecolor{color0}{rgb}{0.12156862745098,0.466666666666667,0.705882352941177}
\definecolor{color1}{rgb}{1,0.498039215686275,0.0549019607843137}
\definecolor{color2}{rgb}{0.172549019607843,0.627450980392157,0.172549019607843}
\scriptsize
\begin{axis}[
width=1.15\linewidth,
title={Memory requirement (MiB)},
legend cell align={left},
legend style={
  fill opacity=0.7,
  draw opacity=1,
  text opacity=1,
  at={(0.03,0.97)},
  anchor=north west,
  draw=white!80!black
},
tick align=outside,
tick pos=left,
xmin=500, xmax=9500,
xtick style={color=black},
ymin=-269.7919232, ymax=5756.2267392,
ytick style={color=black}
]
\addplot [line width=2pt, color0]
table {%
500 379.506688
1000 424.04352
1500 512.263168
2000 625.652736
2500 746.612224
3000 898.689024
3500 1081.34144
4000 1305.76896
4500 1542.347264
5000 1807.367168
5500 2101.377536
6000 2428.296704
6500 2779.136512
7000 3144.467968
7500 3556.375552
8000 3993.277952
8500 4468.419584
9000 4956.025856
9500 5482.3168
};
\addlegendentry{cDGM}
\addplot [line width=2pt, color1]
table {%
500 10.970112
1000 39.070208
1500 83.239936
2000 152.590336
2500 229.004288
3000 326.65088
3500 443.790848
4000 588.02944
4500 739.163648
5000 909.170176
5500 1098.051584
6000 1305.806848
6500 1532.438016
7000 1767.753728
7500 2028.893184
8000 2308.026368
8500 2608.828416
9000 2920.305152
9500 3253.444608
};
\addlegendentry{dDGM}
\addplot [line width=2pt, color2]
table {%
500 4.118016
1000 6.409728
1500 8.705024
2000 10.997248
2500 13.292544
3000 15.586816
3500 17.878016
4000 20.246528
4500 22.46656
5000 24.86016
5500 27.055104
6000 29.348352
6500 31.6416
7000 33.974784
7500 36.884992
8000 39.625216
8500 42.260992
9000 44.824576
9500 46.40512
};
\addlegendentry{dDGM Keops}
\end{axis}

\end{tikzpicture}
        \vspace{-1.5em}
    \end{minipage}
    \caption{\label{fig:speedmemory} Time per iteration and memory requirement for training three different implementations of DGM: continuous graph sampling (cDGM), 5-nn discrete graph sampling implemented with classical Pytorch tensor operations (dDGM) and using the Keops library (dDGM Keops).}
\end{figure}

\begin{table}[b]
    \caption{
    Accuracy on Tadpole transductive task with different architectural choices for our cDGM and dDGM models. The notation $f$+$g$ refers to the choice of the DGM and Diffusion modules (GC: graph convolution, EC: edge convolution; 
    MLP: multilayer perceptron)}. 
    \label{tab:arch}

    \centering
    \begin{tabular}{l | c c c c}
    \hline
         & MLP + GC & GC + GC & MLP + EC & GC + EC \\
         \hline
        cDGM  & 92.42$\pm$3.82& 90.68$\pm$4.58 & 92.29$\pm$4.18 & 91.78$\pm$3.21 \\ 
        dDGM  & 93.47$\pm$3.82 & \textbf{94.09$\pm$1.81} & 93.27$\pm$3.20 & \textbf{94.14$\pm$2.12}  \\ 
        \hline         
    \end{tabular}
\end{table}

\subsection{Application to Healthcare and Brain imaging}
In this section, we show experiments on two medical datasets for disease and age prediction. The main motivation behind the age prediction task is that the difference between brain age and chronicle age is an important bio-marker for identifying certain neurological disorders \cite{niu2020improved}. 

\paragraph*{\textbf{Dataset description}} We use two datasets: {\em Tadpole} \cite{marinescu2018tadpole} comprises multimodal data of 
564 patients, each represented by a 354-dimensional representation derived from imaging (MRI, fMRI, PET) and non-imaging (demographics and genotypes) features. The task is to classify each subject into three classes: `Normal Control', `Alzheimer's Disease' and `Mild Cognitive Impairment'. 
%
\textit{UK Biobank} \cite{miller2016multimodal} comprises multimodal data of 14,503 individuals, each represented by a 440 dimensional feature vector derived from brain MRI and fMRI imaging. 
The task is to classify the age group of the patient (50-59,60-69, 70-79, and 80-89). 
%
We perform the two classification tasks both in a transductive and inductive manner, where in the former setting all the nodes are given during training but the labels of the test nodes are withheld, while in the latter setting, test nodes are completely removed from the population during training, and reintroduced only during testing. 

\paragraph*{\textbf{Architecture}}
For medical applications, the architecture backbone is composed by two diffusion layers with output size of 16 and a final linear layer of size 16 for classification. Each layer is composed by a DGM block and one convolutional layer for diffusion. Since an input graph is not provided with the dataset, we always use a MLP $(d,16,8)$ as the the first DGM graph embedding function. We study the impact of using different operators for the second DGM block and the diffusion convolutions in Table \ref{tab:arch}, showing accuracy performance on the transductive setting for Tadpole. Following these results, we employ Graph Convolution for the second DGM block and Edge Convolution for the diffusion blocks. 
%

\begin{table}[t]
    
    \caption{Classification accuracy in \% on Tadpole in the transductive setting. 
    The top three performance scores are highlighted in color as: {\bf \bf \color{red} First}, {\bf \bf \color{violet} Second}, {\bf Third.}
    }
    \label{Baselines}

    \centering
    \begin{tabular}[]{l c}
    \hline
        Method&Accuracy\\
         \hline
        Linear classifier  & 70.22$\pm$6.32\\
       Multi-GCN \cite{kazi2019self} &76.06$\pm$0.72\\
        Spectral-GCN \cite{parisot2017spectral} &81.00$\pm$6.40 \\
        InceptionGCN \cite{kazi2019inceptiongcn} &84.11$\pm$4.50\\
            DGCNN \cite{wang2019dynamic} & 84.59$\pm$4.33\\
        LDS \cite{franceschi2019learning} &{\bf 87.06$\pm$3.67} \\
        {\bf cDGM}&\textcolor{violet}{\bf 92.91$\pm$2.50}\\
        {\bf dDGM} &\textcolor{red}{\bf 94.14$\pm$2.12}\\
        \hline         
    \end{tabular}
\end{table}

\begin{table}[t]
    \caption{Classification accuracy in \% for disease and age prediction tasks in the \textit{transductive} and \textit{inductive} settings on the Tadpole (left) and UK Biobank datasets (right). $^\dagger$Does not support inductive setting. 
    }
    
    \centering
    \label{ukbb_gender}
        \centering
        
        \begin{tabular}{l|cc|cc}
        \hline
         \multirow{2}{*}{Method} & \multicolumn{2}{c|}{TADPOLE} & \multicolumn{2}{c}{UK Biobank}\\
        \cline{2-5}
        &Transduct.&Inductive&Transduct.&Inductive\\
        \hline
        DGCNN&84.59$\pm$4.33&{\bf 82.99$\pm$4.91} & 58.35$\pm$0.91 & {\bf 51.84$\pm$8.16} \\
        LDS & \textbf{87.06$\pm$3.67} & $\dagger$ & OOM & $\dagger$ \\
        {\bf cDGM} & \textbf{\textcolor{violet}{92.91$\pm$2.50}} & \textbf{\textcolor{violet}{91.85$\pm$2.62}} & \textbf{\textcolor{violet}{61.32$\pm$1.51}} & \textbf{\textcolor{violet}{55.91$\pm$3.49}} \\
        {\bf dDGM} & \textbf{\textcolor{red}{94.10$\pm$2.12}} & \textbf{\textcolor{red}{92.17$\pm$3.65}} & \textbf{ \textcolor{red}{63.22$\pm$1.12} } & \textbf{\textcolor{red}{ 57.34$\pm$5.32}} \\
        \hline
        \end{tabular}
\end{table}

\begin{table}[b]
    \caption{Scalability: training and test iteration times for different number of nodes.
    }
    \label{tab:times}
    \centering
    \begin{tabular}{lccc}
    \hline
%
%
    \multicolumn{4}{c}{Training iteration}\\
    \hline
    Method & $n=$564 & 5k & 10k \\ \hline
    DGCNN & 6.99ms  & 28.2ms & 104ms \\
    LDS \cite{franceschi2019learning} & 1.84s & $>$30m & $>$30m \\
    {\bf cDGM} & 7.35ms & 47.8ms & 211ms \\
    {\bf dDGM} & 8.29ms & 37.0ms & 141ms  \\
    
    \hline\\    
    \hline
    \multicolumn{4}{c}{Test iteration}\\
    \hline
    Method & $n=$564 & 5k & 10k \\ \hline
    DGCNN & 4.60ms  & 25.2ms & 102ms \\
    LDS \cite{franceschi2019learning} & 1.84s & $>$30m & $>$30m \\
    {\bf cDGM} & 3.02ms & 15.1ms & 51ms \\
    {\bf dDGM} & 3.97ms & 24.6ms & 104ms  \\
    \hline
    \end{tabular}
    \vfill
\end{table}

\paragraph*{\textbf{Comparison}}
Previous methods \cite{parisot2017spectral, kazi2019self, kazi2019inceptiongcn,kazi2019graph} used GNNs with hand-crafted patient population graphs built from non-imaging meta-features like the age and sex of patients. 
Our method allows to learn the graph directly from the input patients features, avoiding any handcrafted construction. 
%
We use the following baselines: simple linear classifier as a non-graph method; 
Multi-GCN \cite{kazi2019self}, Spectral-GCN \cite{parisot2017spectral}, InceptionGCN \cite{kazi2019inceptiongcn} as graph methods with hand-crafted graph; and DGCNN \cite{wang2019dynamic} and LDS \cite{franceschi2019learning} as methods that learn the graph. We note that LDS does not support inductive learning. 
%

Results are reported in Tables~\ref{Baselines} and~\ref{ukbb_gender} using 10-fold cross validation. Our model significantly outperforms the state-of-the-art on all the tasks. 
Table~\ref{tab:times} reports the training and testing times of our model for different dataset size, which shows that it is on par with DGCNN and about three orders of magnitude faster than LDS.

\begin{figure*}[t]
    \centering
    \includegraphics[width=0.99\linewidth]{images/shapenet_pbig.pdf}
    \caption{Comparison between our dDGM and DGCNN sampling on the feature space in the last two convolutional layers of the network. In dDGM the colormap encodes the probability of each point to be connected to the red point. For DGCNN we plot the exponential of the negative Euclidean distance on feature space.}
    \label{fig:shapenet_examples} 
\end{figure*}

\subsection{Application to computer vision and Computer graphics}
We also show results of our method on two further domains and applications, namely point cloud segmentation and zero shot learning.

\textbf{Dataset description:} We perform the segmentation of 3D point clouds following \cite{wang2019dynamic} with the same data split. We use 16,881 point clouds from ShapeNet \cite{yi2016scalable}. Each point cloud is sampled at 2048 points, and each point is annotated with one of the 50 part category labels.
%
%
We mimic the DGCNN architecture used in \cite{wang2019dynamic}, replacing their graph kNN sampling scheme by our DGM with a feature depth of 16. We keep the remainder of the network architecture the same, including the value of $k=20$ and training parameters. During inference, given the stochastic nature of our graph, we repeat the classification of each point for 8 times and choose the $argmax$ of the cumulative soft predictions.


Table~\ref{tab:shapenet} reports the mean Intersection-over-Union (mIoU) values calculated by averaging the IoUs of all testing shapes. Our approach allows to increase the performance over the original kNN sampling scheme on almost all shape classes, which is considered very hard.
Figure~\ref{fig:shapenet_examples} depicts the sampling probabilities of some points (denoted by red) on different shapes at the last two layers of the network. We can notice that the probability of connecting two points is not related to the point feature space which is used for part classification, but it rather retains some spatial information and seems to be inspecting symmetries of the shape.


\begin{table}[t]
    \centering
    \caption{Comparison of mIoU(\%) score in ShapeNet part segmentation task. It can be observed in most of the classes dDGM performs better. Since the dataset size is quite large, the overall performance is significantly better.}
    \label{tab:shapenet}    
    \begin{tabular}{l c c c}
        \hline
        & \# Shapes  & DGCNN     &   dDGM \\
        \hline
        Airplane   & 2690 &  84.0          & \textbf{84.1}   \\
        Bag        & 76   &  \textbf{83.4} & 82.5  \\
        Cap        & 55   &  \textbf{86.7} & 84.6  \\
        Car        & 898  &  77.8          & \textbf{77.9}   \\
        Chair      & 3758 &  90.6          & \textbf{91.3}   \\
        Earphone   & 69   &  74.7          & \textbf{79.0}  \\
        Guitar     & 787  &  91.2          & \textbf{92.5}  \\
        Knife      & 392  &  87.5          & \textbf{87.7}  \\
        Lamp       & 1547 &  82.8          & \textbf{83.7} \\
        Laptop     & 451  &  95.7          & \textbf{96.5}   \\
        Motorbike  & 202  &  66.3          & \textbf{66.8}   \\
        Mug        & 184  &  94.9          & \textbf{95.1}   \\
        Pistol     & 283  &  81.1          & \textbf{83.1}  \\
        Rocket     & 66   & \textbf{63.5}  & 62.3  \\
        Skateboard & 152  &  74.5          & \textbf{77.8}  \\
        Table      & 5271 & \textbf{82.6}  & 82.2 \\
        \hline
        MEAN     & & 85.2           & \textbf{85.6}
        \end{tabular}
\end{table}

\begin{figure*}[t]
 \center
  \includegraphics[width=0.352\linewidth]{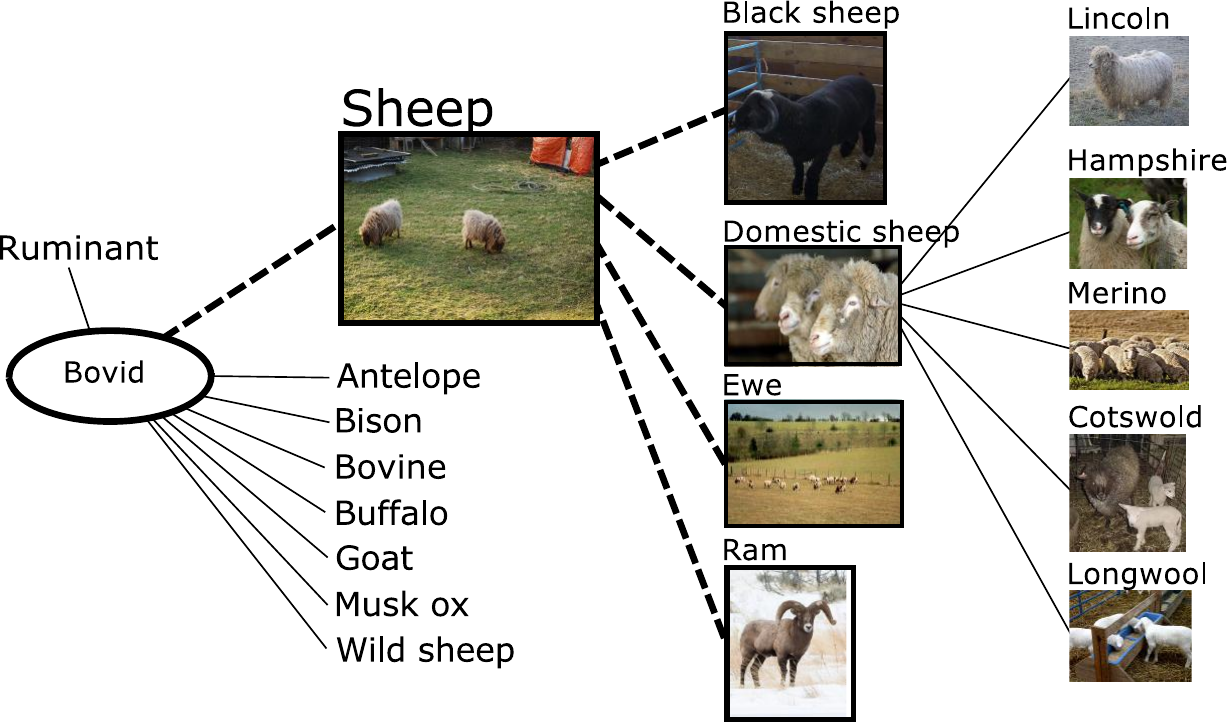}
  \hfill
  \includegraphics[width=0.297\linewidth]{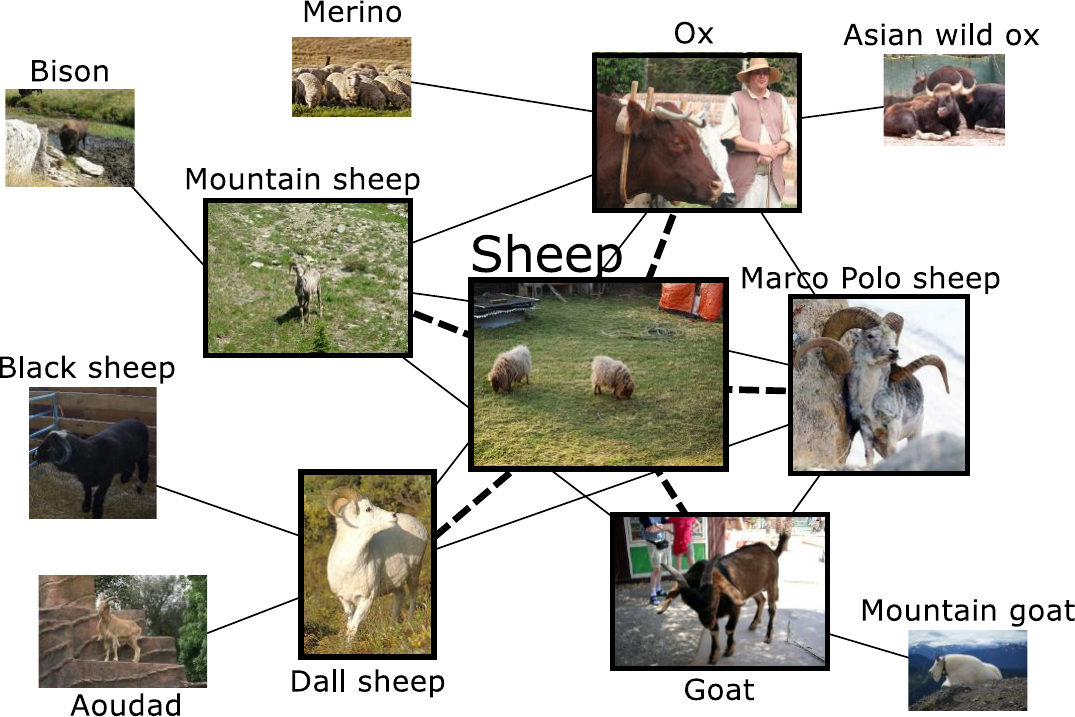}
  \hfill
  \includegraphics[width=0.243\linewidth]{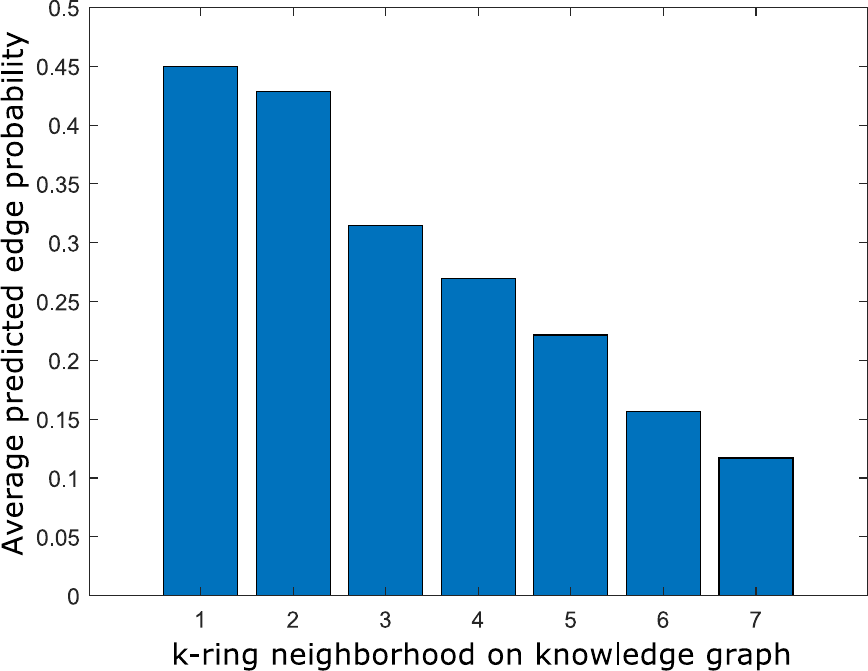}
\caption{Example of the 2-ring neighborhood of the "sheep" category on the knowledge graph (left) and on our predicted graph (center) sampled considering the 5 most probable edges. On the right the average predicted probability of edges belonging to the k-ring neighborhood (AwA2 test categories). Higher probabilities corresponding to nearest neighbors suggest that the predicted graph structure is loosely related to the knowledge graph.}
\label{fig:zeroshot}
\end{figure*}

\subsubsection{Zero-shot learning (ZSL) in computer vision}
%
In ZSL, the task is to learn classifiers for the unseen classes and solve the classification problem for samples belonging to unseen classes based on training data of only seen classes. The most popular approach is to train a network to predict a vector representation for a class starting from some implicit prior knowledge, i.e. semantic embedding \cite{xian2018zero}. Recent works showed that using additional explicit relations between classes in the form of knowledge graphs can help to significantly improve the learning of classifier for the unknown classes and hence the classification accuracy for unseen data samples.
Notably, their method is explicitly devised for using a knowledge graph.  

%
Formally, let $\mathbf{X} \in \mathbb{R}^{N \times S}$ be 
the semantic embeddings (i.e. word vectors) associated with each category class. The Zero-Shot task loss is defined as the summation over all the $M<N$ training classes of $\sum_{i=1}^M \| \mathbf{w}_i - \mathbf{\tilde{w}}_i \|_2^2$, 
where 
$\mathbf{w}_i$ and $\mathbf{\tilde{w}}_i$  are the predicted and ground-truth vector representation of the $i$th class, respectively.
Note that even though we deal with a regression problem in ZSL, it is straightforward to adapt it to deal with our graph loss defined in equation \ref{eq:graph_loss}, considering $\mathrm{argmin}_j \| \mathbf{w}_i-\mathbf{w}_j\|_2$ as the predicted category for sample $\mathbf{x}_i$.

Mimicking \cite{kampffmeyer2019rethinking}, our model consists of two graph convolution layers with hidden and output layer of dimension 2048 and 2049, paired with two DGM layers of dimension 16 for graph representation and $k=3$.
Each layer is composed by the following convolution on graphs:
\begin{equation}
    \mathbf{X}^{(l+1)} = \sigma\left ( (\mathbf{D}^{(l)})^{-1}\mathbf{A}^{(l)} \mathbf{X}^{(l)} \mathbf{\Theta}^{(l)} \right )
    \label{eq:SGCN}
\end{equation}
where $\sigma(\cdot)$ is a \textit{LeakyReLU} non linearity, $\boldsymbol{\Theta}^{(l)}$ are the learned weights and $(\mathbf{D}^{(l)})^{-1}\mathbf{A}^{(l)}$, with $d^{(l)}_{ii}=\sum_j a^{(l)}_{ij}$ is the non-symmetric normalization of the adjacency matrix $\mathbf{A}^{(l)}$. Essentially, our model is the same as SGCN \cite{kampffmeyer2019rethinking}, where we replace the input knowledge graph by our DGM module for learning $\mathbf{A}$.
%


Following \cite{kampffmeyer2019rethinking}, we use weights of the last fully connected layer of a ResNet-50 \cite{he2016deep} (pre-trained on ImageNet 2012 dataset \cite{deng2009imagenet}) as our target vector representation $\mathbf{\tilde{y}}_i \in \mathbb{R}^{2049}$. Input semantic features $\mathbf{x}_i \in \mathbb{R}^{300}$ are extracted with the GloVe text model \cite{pennington2014glove} (trained on Wikipedia dataset). 

We train our model on the 21K ImageNet dataset classes, where we have the semantic embedding for all classes as input, but only the first 1K have a corresponding ground-truth vector representation.
The model is trained for 5000 iterations on a randomly subsampled set of 7K categories containing all the 1K training samples.

Testing is performed on the AWA2 dataset, composed of 37,322 images belonging to 50 different animal classes. We used the test split proposed in \cite{wang2018zero}, comprising images from 10 classes not present in the first 1K training samples of ImageNet.  
The Top-1 accuracy for dDGM is $74.7\%$, which is reater than that of GCNZ \cite{wang2018zero} ($70.5\%$), but lower than DGP ($77.3\%$). Note that compared to the latter two methods, our dDGM approach does not make use of the knowledge graph. As shown in Figure \ref{fig:zeroshot}, the knowledge graph indeed seems to be a good graph representation for the zero-shot task. Our predicted graph shows some similarity to it, however, fails to capture its hierarchical structure. We leave it for future work to impose additional constraints on the graph structure, e.g. encouraging a more tree-like structure.

\section{Discussion and conclusion}
%
In this paper, we tackled the challenge of graph learning in convolutional graph neural networks. We have proposed a novel Differentiable Graph Module (DGM): it predicts a probabilistic graph, from which a discrete graph can be sampled, which can then be used in any graph convolutional operator towards the downstream task. Furthermore, we devised a weighted loss 
that models the optimization over edge probabilities.

Our DGM is generic and adaptable to any graph convolution based method. We prove this by using our method to solve a wide variety of tasks, starting from applications in healthcare (disease prediction), brain imaging (age and gender prediction), computer graphics (3D point cloud segmentation) and computer vision (zero-shot learning). These applications feature both multimodal datasets and inductive inference setups. 

There are several avenues for future work in our proposed method. 
For example, despite being computationally more lightweight than existing approaches (e.g. \cite{soobeom2019}), our method still has a quadratic complexity with respect to the number of input nodes, as it requires the computation of all pairwise distances. Restricting the computation of probabilities in a neighborhood of the node and using tree-based algorithms could help in reducing the complexity to $\mathcal{O}(n\log{}n)$.
Further, our choice of sampling $k$ neighbors does not consider the heterogeneity of the graph in terms of the degree distribution of nodes. Other sampling schemes (e.g. threshold-based sampling \cite{soobeom2019}) could be investigated. It would be also interesting to consider previous knowledge about the graph, e.g. by imposing a node degree distribution, or providing an initial input graph to be optimized for a specific task.






%


\ifCLASSOPTIONcompsoc
  \section*{Acknowledgments}
\else
  \section*{Acknowledgment}
\fi
The study was carried out with financial support of TUM-ICL incentive funding, Freunde und F{\"o}rderer der Augenklinik, M{\"u}nchen, Germany and ERC Consolidator grant No. 724228 (LEMAN) and German Federal Ministry of Education and Health (BMBF) in connection with the foundation of the German Center for Vertigo and Balance Disorders (DSGZ) [grant number 01 EO0901]. The UK Biobank data is used under the application id 51541.\\




\bibliographystyle{IEEEtran}
\bibliography{IEEEabrv,bibliography}

\begin{thebibliography}{10}
\providecommand{\url}[1]{#1}
\csname url@samestyle\endcsname
\providecommand{\newblock}{\relax}
\providecommand{\bibinfo}[2]{#2}
\providecommand{\BIBentrySTDinterwordspacing}{\spaceskip=0pt\relax}
\providecommand{\BIBentryALTinterwordstretchfactor}{4}
\providecommand{\BIBentryALTinterwordspacing}{\spaceskip=\fontdimen2\font plus
\BIBentryALTinterwordstretchfactor\fontdimen3\font minus
  \fontdimen4\font\relax}
\providecommand{\BIBforeignlanguage}[2]{{%
\expandafter\ifx\csname l@#1\endcsname\relax
\typeout{** WARNING: IEEEtran.bst: No hyphenation pattern has been}%
\typeout{** loaded for the language `#1'. Using the pattern for}%
\typeout{** the default language instead.}%
\else
\language=\csname l@#1\endcsname
\fi
#2}}
\providecommand{\BIBdecl}{\relax}
\BIBdecl

\bibitem{bronstein2017geometric}
M.~M. Bronstein, J.~Bruna, Y.~LeCun, A.~Szlam, and P.~Vandergheynst,
  ``Geometric deep learning: going beyond euclidean data,'' \emph{IEEE Signal
  Process. Mag.}, vol.~34, no.~4, pp. 18--42, 2017.

\bibitem{hamilton2017representation}
W.~L. Hamilton, R.~Ying, and J.~Leskovec, ``Representation learning on graphs:
  Methods and applications,'' \emph{arXiv:1709.05584}, 2017.

\bibitem{battaglia2018relational}
P.~W. Battaglia \emph{et~al.}, ``Relational inductive biases, deep learning,
  and graph networks,'' \emph{arXiv:1806.01261}, 2018.

\bibitem{zhang2018link}
M.~Zhang and Y.~Chen, ``Link prediction based on graph neural networks,'' in
  \emph{Proc. NeurIPS}, 2018.

\bibitem{qi2018learning}
S.~Qi, W.~Wang, B.~Jia, J.~Shen, and S.-C. Zhu, ``Learning human-object
  interactions by graph parsing neural networks,'' in \emph{ECCV}, 2018, pp.
  401--417.

\bibitem{qi20173d}
X.~Qi, R.~Liao, J.~Jia, S.~Fidler, and R.~Urtasun, ``3d graph neural networks
  for rgbd semantic segmentation,'' in \emph{ICCV}, 2017, pp. 5199--5208.

\bibitem{monti2017geometric}
F.~Monti, D.~Boscaini, J.~Masci, E.~Rodola, J.~Svoboda, and M.~M. Bronstein,
  ``Geometric deep learning on graphs and manifolds using mixture model cnns,''
  in \emph{Proc. CVPR}, 2017.

\bibitem{wang2019dynamic}
Y.~Wang, Y.~Sun, Z.~Liu, S.~E. Sarma, M.~M. Bronstein, and J.~M. Solomon,
  ``Dynamic graph cnn for learning on point clouds,'' \emph{ACM TOG}, vol.~38,
  no.~5, p. 146, 2019.

\bibitem{choma2018graph}
N.~Choma \emph{et~al.}, ``Graph neural networks for icecube signal
  classification,'' in \emph{Proc. ICMLA}, 2018.

\bibitem{duvenaud2015convolutional}
D.~K. Duvenaud, D.~Maclaurin, J.~Iparraguirre, R.~Bombarell, T.~Hirzel,
  A.~Aspuru-Guzik, and R.~P. Adams, ``Convolutional networks on graphs for
  learning molecular fingerprints,'' in \emph{NeurIPS}, 2015, pp. 2224--2232.

\bibitem{gilmer2017neural}
J.~Gilmer, S.~S. Schoenholz, P.~F. Riley, O.~Vinyals, and G.~E. Dahl, ``Neural
  message passing for quantum chemistry,'' in \emph{International conference on
  machine learning}.\hskip 1em plus 0.5em minus 0.4em\relax PMLR, 2017, pp.
  1263--1272.

\bibitem{li2018learning}
Y.~Li, O.~Vinyals, C.~Dyer, R.~Pascanu, and P.~Battaglia, ``Learning deep
  generative models of graphs,'' \emph{arXiv preprint arXiv:1803.03324}, 2018.

\bibitem{parisot2018disease}
S.~Parisot, S.~I. Ktena, E.~Ferrante, M.~Lee, R.~Guerrero, B.~Glocker, and
  D.~Rueckert, ``Disease prediction using graph convolutional networks:
  Application to autism spectrum disorder and alzheimer’s disease,''
  \emph{Med Image Anal}, vol.~48, pp. 117--130, 2018.

\bibitem{parisot2017spectral}
S.~Parisot, S.~I. Ktena, E.~Ferrante, M.~Lee, R.~G. Moreno, B.~Glocker, and
  D.~Rueckert, ``Spectral graph convolutions for population-based disease
  prediction,'' in \emph{MICCAI}.\hskip 1em plus 0.5em minus 0.4em\relax
  Springer, 2017, pp. 177--185.

\bibitem{mellema2019multiple}
C.~Mellema, A.~Treacher, K.~Nguyen, and A.~Montillo, ``Multiple deep learning
  architectures achieve superior performance diagnosing autism spectrum
  disorder using features previously extracted from structural and functional
  mri,'' in \emph{2019 IEEE ISBI)}.\hskip 1em plus 0.5em minus 0.4em\relax
  IEEE, 2019, pp. 1891--1895.

\bibitem{kazi2019inceptiongcn}
A.~Kazi, S.~Shekarforoush, S.~A. Krishna, H.~Burwinkel, G.~Vivar,
  K.~Kort{\"u}m, S.-A. Ahmadi, S.~Albarqouni, and N.~Navab, ``Inceptiongcn:
  Receptive field aware graph convolutional network for disease prediction,''
  in \emph{IPMI}.\hskip 1em plus 0.5em minus 0.4em\relax Springer, 2019.

\bibitem{zitnik2018modeling}
M.~Zitnik, M.~Agrawal, and J.~Leskovec, ``Modeling polypharmacy side effects
  with graph convolutional networks,'' \emph{Bioinformatics}, vol.~34, no.~13,
  pp. i457--i466, 2018.

\bibitem{Zitnik19}
M.~Zitnik, F.~Nguyen, B.~Wang, J.~Leskovec, A.~Goldenberg, and M.~M. Hoffman,
  ``Machine learning for integrating data in biology and medicine: Principles,
  practice, and opportunities,'' \emph{Information Fusion}, vol.~50, pp.
  71--91, 2019.

\bibitem{gainza2019deciphering}
P.~Gainza, F.~Sverrisson, F.~Monti, E.~Rodola, M.~M. Bronstein, and B.~E.
  Correia, ``Deciphering interaction fingerprints from protein molecular
  surfaces using geometric deep learning,'' \emph{Nature Methods}, vol.~17, pp.
  184--192, 2019.

\bibitem{scarselli2008graph}
F.~Scarselli, M.~Gori, A.~C. Tsoi, M.~Hagenbuchner, and G.~Monfardini, ``The
  graph neural network model,'' \emph{IEEE Trans. Neural Netw}, vol.~20, no.~1,
  pp. 61--80, 2008.

\bibitem{bruna2013spectral}
J.~Bruna, W.~Zaremba, A.~Szlam, and Y.~LeCun, ``Spectral networks and locally
  connected networks on graphs,'' \emph{arXiv preprint arXiv:1312.6203}, 2013.

\bibitem{defferrard2016convolutional}
M.~Defferrard, X.~Bresson, and P.~Vandergheynst, ``Convolutional neural
  networks on graphs with fast localized spectral filtering,'' in
  \emph{NeurIPS}, 2016, pp. 3844--3852.

\bibitem{kipf2016semi}
T.~N. Kipf and M.~Welling, ``Semi-supervised classification with graph
  convolutional networks,'' \emph{arXiv preprint arXiv:1609.02907}, 2016.

\bibitem{levie2018cayleynets}
R.~Levie, F.~Monti, X.~Bresson, and M.~M. Bronstein, ``Cayleynets: Graph
  convolutional neural networks with complex rational spectral filters,''
  \emph{IEEE Trans. Signal Process.}, vol.~67, no.~1, pp. 97--109, 2018.

\bibitem{bianchi2019graph}
F.~M. Bianchi, D.~Grattarola, C.~Alippi, and L.~Livi, ``Graph neural networks
  with convolutional arma filters,'' \emph{arXiv:1901.01343}, 2019.

\bibitem{velivckovic2017graph}
P.~Veli{\v{c}}kovi{\'c}, G.~Cucurull, A.~Casanova, A.~Romero, P.~Lio, and
  Y.~Bengio, ``Graph attention networks,'' \emph{arXiv:1710.10903}, 2017.

\bibitem{kondor2018n}
R.~Kondor, ``N-body networks: a covariant hierarchical neural network
  architecture for learning atomic potentials,'' \emph{arXiv:1803.01588}, 2018.

\bibitem{bruna2017community}
J.~Bruna and X.~Li, ``Community detection with graph neural networks,''
  \emph{Stat}, vol. 1050, p.~27, 2017.

\bibitem{monti2018dual}
F.~Monti \emph{et~al.}, ``Dual-primal graph convolutional networks,''
  \emph{arXiv:1806.00770}, 2018.

\bibitem{hamilton2017inductive}
W.~Hamilton, Z.~Ying, and J.~Leskovec, ``Inductive representation learning on
  large graphs,'' in \emph{Proc. NIPS}, 2017.

\bibitem{pinsage}
R.~Ying, R.~He, K.~Chen, P.~Eksombatchai, W.~L. Hamilton, and J.~Leskovec,
  ``Graph convolutional neural networks for web-scale recommender systems,'' in
  \emph{KDD}, 2018.

\bibitem{Chiang:2019:CEA:3292500.3330925}
W.-L. Chiang, X.~Liu, S.~Si, Y.~Li, S.~Bengio, and C.-J. Hsieh, ``Cluster-gcn:
  An efficient algorithm for training deep and large graph convolutional
  networks,'' in \emph{KDD}, 2019.

\bibitem{DBLP:journals/corr/abs-1907-04931}
H.~Zeng, H.~Zhou, A.~Srivastava, R.~Kannan, and V.~K. Prasanna, ``Graphsaint:
  Graph sampling based inductive learning method,'' \emph{arXiv:1907.04931},
  2019.

\bibitem{rong2019dropedge}
Y.~Rong, W.~Huang, T.~Xu, and J.~Huang, ``Dropedge: Towards deep graph
  convolutional networks on node classification,'' \emph{arXiv preprint
  arXiv:1907.10903}, 2019.

\bibitem{zhao2019pairnorm}
L.~Zhao and L.~Akoglu, ``Pairnorm: Tackling oversmoothing in gnns,'' in
  \emph{Proc. ICLR}, 2020.

\bibitem{li2019deepgcns}
G.~Li, M.~Muller, A.~Thabet, and B.~Ghanem, ``Deepgcns: Can gcns go as deep as
  cnns?'' in \emph{Proc. ICCV}, 2019.

\bibitem{gong2020geometrically}
S.~Gong, M.~Bahri, M.~M. Bronstein, and S.~Zafeiriou, ``Geometrically
  principled connections in graph neural networks,'' in \emph{Proc. CVPR},
  2020.

\bibitem{xu2018how}
K.~Xu, W.~Hu, J.~Leskovec, and S.~Jegelka, ``How powerful are graph neural
  networks?'' in \emph{International Conference on Learning Representations
  (ICLR)}, 2019.

\bibitem{DBLP:conf/iclr/MaronBSL19}
H.~Maron, H.~Ben{-}Hamu, N.~Shamir, and Y.~Lipman, ``Invariant and equivariant
  graph networks,'' in \emph{International Conference on Learning
  Representations, {ICLR}}, 2019.

\bibitem{keriven2019universal}
N.~Keriven and G.~Peyr{\'e}, ``Universal invariant and equivariant graph neural
  networks,'' in \emph{Advances in Neural Information Processing Systems},
  2019, pp. 7090--7099.

\bibitem{Loukas2020What}
A.~Loukas, ``What graph neural networks cannot learn: depth vs width,'' in
  \emph{International Conference on Learning Representations (ICLR)}, 2020.

\bibitem{liu2012robust}
W.~Liu, J.~Wang, and S.-F. Chang, ``Robust and scalable graph-based
  semisupervised learning,'' \emph{Proc. IEEE}, vol. 100, no.~9, pp.
  2624--2638, 2012.

\bibitem{Dong19}
X.~Dong, D.~Thanou, M.~Rabbat, and P.~Frossard, ``Learning graphs from data: A
  signal representation perspective,'' \emph{IEEE Signal Processing Magazine},
  vol.~36, no.~3, pp. 44--63, 2019.

\bibitem{Mateos19}
G.~Mateos, S.~Segarra, A.~G. Marques, and A.~Ribeiro, ``Connecting the dots:
  Identifying network structure via graph signal processing,'' \emph{IEEE
  Signal Processing Magazine}, vol.~36, no.~3, pp. 16--43, 2019.

\bibitem{cosmo2020latent}
C.~et~al., ``Latent-graph learning for disease prediction,'' in
  \emph{MICCAI}.\hskip 1em plus 0.5em minus 0.4em\relax Springer, 2020.

\bibitem{dong2019learning}
X.~Dong, D.~Thanou, M.~Rabbat, and P.~Frossard, ``Learning graphs from data: A
  signal representation perspective,'' \emph{IEEE Signal Processing Magazine},
  vol.~36, no.~3, pp. 44--63, 2019.

\bibitem{kipf2018neural}
T.~Kipf, E.~Fetaya, K.-C. Wang, M.~Welling, and R.~Zemel, ``Neural relational
  inference for interacting systems,'' \emph{arXiv preprint arXiv:1802.04687},
  2018.

\bibitem{zhang2018gaan}
J.~Zhang, X.~Shi, J.~Xie, H.~Ma, I.~King, and D.-Y. Yeung, ``Gaan: Gated
  attention networks for learning on large and spatiotemporal graphs,''
  \emph{arXiv preprint arXiv:1803.07294}, 2018.

\bibitem{lee2018graph}
J.~B. Lee, R.~Rossi, and X.~Kong, ``Graph classification using structural
  attention,'' in \emph{Proceedings of the 24th ACM SIGKDD International
  Conference on Knowledge Discovery \& Data Mining}, 2018, pp. 1666--1674.

\bibitem{abu2017watch}
S.~Abu-El-Haija, B.~Perozzi, R.~Al-Rfou, and A.~Alemi, ``Watch your step:
  Learning node embeddings via graph attention,'' \emph{arXiv preprint
  arXiv:1710.09599}, 2017.

\bibitem{liu2019geniepath}
Z.~Liu, C.~Chen, L.~Li, J.~Zhou, X.~Li, L.~Song, and Y.~Qi, ``Geniepath: Graph
  neural networks with adaptive receptive paths,'' in \emph{Proceedings of the
  AAAI Conference on Artificial Intelligence}, vol.~33, no.~01, 2019, pp.
  4424--4431.

\bibitem{tran2018filter}
D.~V. Tran, N.~Navarin, and A.~Sperduti, ``On filter size in graph
  convolutional networks,'' in \emph{2018 IEEE Symposium Series on
  Computational Intelligence (SSCI)}.\hskip 1em plus 0.5em minus 0.4em\relax
  IEEE, 2018, pp. 1534--1541.

\bibitem{zhan2018adaptive}
K.~Zhan, X.~Chang, J.~Guan, L.~Chen, Z.~Ma, and Y.~Yang, ``Adaptive structure
  discovery for multimedia analysis using multiple features,'' \emph{IEEE
  transactions on cybernetics}, vol.~49, no.~5, pp. 1826--1834, 2018.

\bibitem{li2018adaptive}
R.~Li, S.~Wang, F.~Zhu, and J.~Huang, ``Adaptive graph convolutional neural
  networks,'' in \emph{AAAI}, 2018.

\bibitem{huang2018adaptive}
W.~Huang, T.~Zhang, Y.~Rong, and J.~Huang, ``Adaptive sampling towards fast
  graph representation learning,'' in \emph{NeurIPS}, 2018, pp. 4558--4567.

\bibitem{jiang2019semi}
B.~Jiang, Z.~Zhang, D.~Lin, J.~Tang, and B.~Luo, ``Semi-supervised learning
  with graph learning-convolutional networks,'' in \emph{Proceedings of the
  IEEE Conference on Computer Vision and Pattern Recognition}, 2019, pp.
  11\,313--11\,320.

\bibitem{franceschi2019learning}
L.~Franceschi, M.~Niepert, M.~Pontil, and X.~He, ``Learning discrete structures
  for graph neural networks,'' in \emph{International conference on machine
  learning}.\hskip 1em plus 0.5em minus 0.4em\relax PMLR, 2019, pp. 1972--1982.

\bibitem{yang2019modeling}
Y.~et~al., ``Modeling point clouds with self-attention and gumbel subset
  sampling,'' in \emph{CVPR}, 2019.

\bibitem{ying2018hierarchical}
Z.~Ying, J.~You, C.~Morris, X.~Ren, W.~Hamilton, and J.~Leskovec,
  ``Hierarchical graph representation learning with differentiable pooling,''
  \emph{Advances in neural information processing systems}, vol.~31, 2018.

\bibitem{norcliffe2018learning}
N.~et~al., ``Learning conditioned graph structures for interpretable visual
  question answering,'' \emph{arXiv preprint arXiv:1806.07243}, 2018.

\bibitem{chen2020iterative}
Y.~et~al., ``Iterative deep graph learning for graph neural networks:better and
  robust node embeddings,'' \emph{NeurIPS}, 2020.

\bibitem{cosmo2021graph}
L.~Cosmo, G.~Minello, M.~Bronstein, E.~Rodol{\`a}, L.~Rossi, and A.~Torsello,
  ``Graph kernel neural networks,'' \emph{arXiv preprint arXiv:2112.07436},
  2021.

\bibitem{krioukov2010hyperbolic}
D.~Krioukov, F.~Papadopoulos, M.~Kitsak, A.~Vahdat, and M.~Bogun{\'a},
  ``Hyperbolic geometry of complex networks,'' \emph{Physical Review E},
  vol.~82, no.~3, p. 036106, 2010.

\bibitem{poincare_embeddings}
M.~Nickel and D.~Kiela, ``Poincar{\'e} embeddings for learning hierarchical
  representations,'' in \emph{Advances in neural information processing
  systems}, 2017, pp. 6338--6347.

\bibitem{wouter2019}
\BIBentryALTinterwordspacing
W.~Kool, H.~van Hoof, and M.~Welling, ``Stochastic beams and where to find
  them: The gumbel-top-k trick for sampling sequences without replacement,''
  \emph{CoRR}, vol. abs/1903.06059, 2019. [Online]. Available:
  \url{http://arxiv.org/abs/1903.06059}
\BIBentrySTDinterwordspacing

\bibitem{planetoid}
P.~Sen, G.~Namata, M.~Bilgic, L.~Getoor, B.~Galligher, and T.~Eliassi-Rad,
  ``Collective classification in network data,'' \emph{AI Magazine}, vol.~29,
  no.~3, p.~93, Sep. 2008.

\bibitem{smith2019geometry}
A.~L. Smith, D.~M. Asta, and C.~A. Calder, ``The geometry of continuous latent
  space models for network data,'' \emph{Statistical science: a review journal
  of the Institute of Mathematical Statistics}, vol.~34, no.~3, p. 428, 2019.

\bibitem{chami2019hyperbolic}
I.~Chami, R.~Ying, C.~R{\'e}, and J.~Leskovec, ``Hyperbolic graph convolutional
  neural networks,'' \emph{Advances in neural information processing systems},
  vol.~32, p. 4869, 2019.

\bibitem{beltrami1868teoria}
E.~Beltrami, \emph{Teoria fondamentale degli spazii di curvatura costante:
  memoria}.\hskip 1em plus 0.5em minus 0.4em\relax F. Zanetti, 1868.

\bibitem{charlier2020kernel}
B.~Charlier, J.~Feydy, J.~A. Glaun{\`e}s, F.-D. Collin, and G.~Durif, ``Kernel
  operations on the gpu, with autodiff, without memory overflows,'' \emph{arXiv
  preprint arXiv:2004.11127}, 2020.

\bibitem{niu2020improved}
X.~Niu, F.~Zhang, J.~Kounios, and H.~Liang, ``Improved prediction of brain age
  using multimodal neuroimaging data,'' \emph{Human brain mapping}, vol.~41,
  no.~6, pp. 1626--1643, 2020.

\bibitem{marinescu2018tadpole}
R.~V. Marinescu, N.~P. Oxtoby, A.~L. Young, E.~E. Bron, A.~W. Toga, M.~W.
  Weiner, F.~Barkhof, N.~C. Fox, S.~Klein, D.~C. Alexander \emph{et~al.},
  ``Tadpole challenge: Prediction of longitudinal evolution in alzheimer's
  disease,'' \emph{arXiv preprint arXiv:1805.03909}, 2018.

\bibitem{miller2016multimodal}
K.~L. Miller, F.~Alfaro-Almagro, N.~K. Bangerter, D.~L. Thomas, E.~Yacoub,
  J.~Xu, A.~J. Bartsch, S.~Jbabdi, S.~N. Sotiropoulos, J.~L. Andersson
  \emph{et~al.}, ``Multimodal population brain imaging in the uk biobank
  prospective epidemiological study,'' \emph{Nature neuroscience}, vol.~19,
  no.~11, p. 1523, 2016.

\bibitem{kazi2019self}
A.~Kazi, S.~Shekarforoush, K.~Kortuem, S.~Albarqouni, N.~Navab \emph{et~al.},
  ``Self-attention equipped graph convolutions for disease prediction,'' in
  \emph{ISBI}.\hskip 1em plus 0.5em minus 0.4em\relax IEEE, 2019, pp.
  1896--1899.

\bibitem{kazi2019graph}
A.~Kazi, S.~Shekarforoush, S.~A. Krishna, H.~Burwinkel, G.~Vivar, B.~Wiestler,
  K.~Kort{\"u}m, S.-A. Ahmadi, S.~Albarqouni, and N.~Navab, ``Graph convolution
  based attention model for personalized disease prediction,'' in
  \emph{International Conference on Medical Image Computing and
  Computer-Assisted Intervention}.\hskip 1em plus 0.5em minus 0.4em\relax
  Springer, 2019, pp. 122--130.

\bibitem{yi2016scalable}
L.~Yi, V.~G. Kim, D.~Ceylan, I.-C. Shen, M.~Yan, H.~Su, C.~Lu, Q.~Huang,
  A.~Sheffer, and L.~Guibas, ``A scalable active framework for region
  annotation in 3d shape collections,'' \emph{ACM TOG}, vol.~35, no.~6, pp.
  1--12, 2016.

\bibitem{xian2018zero}
Y.~Xian, C.~H. Lampert, B.~Schiele, and Z.~Akata, ``Zero-shot learning-a
  comprehensive evaluation of the good, the bad and the ugly,'' \emph{IEEE
  PAMI}, 2018.

\bibitem{kampffmeyer2019rethinking}
M.~Kampffmeyer, Y.~Chen, X.~Liang, H.~Wang, Y.~Zhang, and E.~P. Xing,
  ``Rethinking knowledge graph propagation for zero-shot learning,'' in
  \emph{CVPR}, 2019, pp. 11\,487--11\,496.

\bibitem{he2016deep}
K.~He, X.~Zhang, S.~Ren, and J.~Sun, ``Deep residual learning for image
  recognition,'' in \emph{CVPR}, 2016, pp. 770--778.

\bibitem{deng2009imagenet}
J.~Deng, W.~Dong, R.~Socher, L.-J. Li, K.~Li, and L.~Fei-Fei, ``Imagenet: A
  large-scale hierarchical image database,'' in \emph{CVPR}.\hskip 1em plus
  0.5em minus 0.4em\relax Ieee, 2009.

\bibitem{pennington2014glove}
J.~Pennington, R.~Socher, and C.~D. Manning, ``Glove: Global vectors for word
  representation,'' in \emph{EMNLP}, 2014, pp. 1532--1543.

\bibitem{wang2018zero}
X.~Wang, Y.~Ye, and A.~Gupta, ``Zero-shot recognition via semantic embeddings
  and knowledge graphs,'' in \emph{CVPR}, 2018, pp. 6857--6866.

\bibitem{soobeom2019}
\BIBentryALTinterwordspacing
S.~Jang, S.~Moon, and J.~Lee, ``Brain signal classification via learning
  connectivity structure,'' \emph{CoRR}, vol. abs/1905.11678, 2019. [Online].
  Available: \url{http://arxiv.org/abs/1905.11678}
\BIBentrySTDinterwordspacing

\end{thebibliography}

%
\newpage

\vfill


\end{document}